\documentclass[journal]{IEEEtran}

\usepackage{epsfig}
\usepackage{epstopdf}

\usepackage{amssymb}
\usepackage{amsmath,amsthm}
\usepackage[ruled,vlined]{algorithm2e}
\usepackage{multirow,booktabs}
\usepackage{graphicx}  
\usepackage{subfigure}
\usepackage{epstopdf}
\usepackage{subfloat}
\usepackage{hyperref}
\usepackage{subfloat}
\usepackage{url}
\usepackage{dsfont}
\usepackage{bm}
\usepackage{cite}

\usepackage{wasysym}
\newcommand{\rknorm}[1]{\left\| #1 \right\|_\mathcal{H}}
\newcommand{\rkinner}[2]{\langle #1, #2 \rangle_\mathcal{H}}
\newcommand{\infnorm}[1]{\left\| #1 \right\|_\infty}
\newcommand{\onenorm}[1]{\left\| #1 \right\|_1}

\DeclareMathOperator*{\argmin}{\arg\,\min}  
\newcommand*{\QED}{\hfill\ensuremath{\square}}%

\newtheorem{theorem}{Theorem}

\newtheorem{proposition}{Proposition}

\newtheorem{myAssumption}{Assumption}

\theoremstyle{definition}
\newtheorem*{myProof}{Proof}
\newtheorem{myRemark}{Remark}

\usepackage{xcolor}
\definecolor{comment}{rgb}{0.0,0.5,0.0}

\newcommand{\red}[1]{{\color{black}#1}}

\SetKwRepeat{Do}{do}{while}

\usepackage{hyperref}

\begin{document}

\title{\LARGE\bf Robust Uncertainty Bounds in Reproducing Kernel Hilbert Spaces: A Convex Optimization Approach}
\author{ Paul Scharnhorst$^*$, Emilio T. Maddalena$^*$, Yuning Jiang, Colin N. Jones
\thanks{*The first two authors contributed equally. This work has received support from the Swiss National Science Foundation under the RISK project (Risk Aware Data-Driven Demand Response, grant number 200021 175627), and CSEM's Data Program. (Corresponding author: Yuning Jiang)}
\thanks{All authors are with Automatic Control Laboratory, EPFL, Lausanne, Switzerland. Paul Scharnhorst is with CSEM, Neuchâtel, Switzerland. 
(e-mail: {\tt paul.scharnhorst, emilio.maddalena, yuning.jiang, colin.jones@epfl.ch}) }
}

\maketitle

\begin{abstract}
	
	\textcolor{black}{
	The problem of establishing out-of-sample bounds for the values of an unkonwn ground-truth function is considered. Kernels and their associated Hilbert spaces are the main formalism employed herein along with an observational model where outputs are corrupted by bounded measurement noise. The noise can originate from any compactly supported distribution and no independence assumptions are made on the available data.
	In this setting, we show how computing tight, finite-sample uncertainty bounds amounts to solving parametric quadratically constrained linear programs. Next, properties of our approach are established and its relationship with another methods is studied. Numerical experiments are presented to exemplify how the theory can be applied in a number of scenarios, and to contrast it with other closed-form alternatives.}
\end{abstract}

\section{Introduction}
\label{sec::Intro}
We consider the problem of quantifying the uncertainty associated with point-evaluations of an unknown ground-truth map given a dataset of observations and assumptions on its nature. The analysis differs from widespread concentration bounds found in the machine learning literature as no assumptions are made on the statistical independence of samples. This agnosticism is central when dealing with systems that incorporate memory such as physical plants that evolve in a dynamical and hence strongly correlated fashion|see \cite{scholkopf2019causality} for a thorough discussion about situations where the typical i.i.d. premise is inadequate. In exchange, we pose conditions on the ground-truth, requiring it to belong to a specific class of functions \cite{cucker2007learning}, and allow for observations to be scattered, not necessarily being drawn from any specific distribution. This is rather customary in the established field of approximation theory \cite{schaback1995error,wendland2004scattered,iske2018approximation}. See also \cite{belkin2018approximation} for a recent perspective on the advantages offered by approximation-type bounds.

The setting in this paper is that of kernel learning, which is among the most prominent modern frameworks for both classification and regression. These non-parametric techniques are usually more data-efficient than deep network architectures, and recently intriguing connections between these two methodologies were established \cite{mei2019mean,domingos2020every,de2018gaussian}. 
Kernels can be regarded as similarity measures between examples in a certain feature space \cite{scholkopf2002learning}. This space is known as the reproducing kernel Hilbert space (RKHS) and is usually an infinite-dimensional linear space of functions. Moreover, it is well known that the RKHS associated with certain kernel classes is dense in the space of continuous functions on compact domains \cite{micchelli2006universal}. By requiring the latent ground-truth to be a member of the RKHS associated with a known kernel, straightforward error-bounds can be established for models that interpolate noise-free data-points (see for instance \cite{schaback2006kernel}). Recently, these out-of-sample guarantees were extended to regularized smoothing models in the presence of bounded measurement noise \cite{maddalena2020deterministic}. Nevertheless, the task of \textit{exactly} quantifying the associated uncertainty in the latter scenario remained open, as well as understanding how much conservativeness is introduced when centering the bounds around pre-specified models.

\textbf{Contributions:} Herein we investigate the uncertainty quantification problem in RKHSs and with datasets corrupted by measurement noise. The sources of uncertainty are both epistemic and aleatoric \cite{o2004dicing} as explained next. The first stems from the ground-truth being an unknown fixed member of our function class, and from which we derive information indirectly through its samples. Secondly, the additive bounded measurement noise, which could originate from any probability measure, or even be a constant, fixed bias. In contrast with the study in \cite{maddalena2020deterministic}, we carry out an \textit{algorithmic independent} analysis that is not centered around any specific model; this is done by computing the highest and lowest possible point-evaluations that are consistent with our knowledge. Our main result is to show how this infinite-dimensional problem can be translated into a finite convex quadratically constrained linear program (QCLP) without any conservatism, which is accomplished through a representer theorem. Next, properties of this procedure are derived and connections with closed-form sub-optimal bounds \cite{maddalena2020deterministic} as well as classical noise-free bounds \cite{wendland2004scattered} are established. Finally, efficient solution methods are proposed through the dual optimization problem, which trade-off computational time and precision. Numerical experiments are reported to illustrate their use, as well as the influence of the input distribution on the final results. 

\textbf{Relevance for automatic control:} The use of the so-called data-driven techniques to refine models, improve performance on-line, or approximate controllers is becoming evermore present in the field of automatic control \cite{chakrabarty2016support,rosolia2017learning,umlauft2019feedback}. In the particular case of kernel surrogate models, a considerable body of rigorous literature exists for linear dynamics \cite{chen2014system,pillonetto2014kernel,pin2015non,fujimoto2018input}, linear parameter-varying dynamics \cite{rizvi2018state,laurain2020sparse}, Hammerstein and Wiener cascaded systems \cite{risuleo2017nonparametric,risuleo2019bayesian}, mainly adopting a time-domain perspective of the identification problem. When operational constraints are present, one has to pair these tools with appropriate uncertainty quantification techniques to not make unsafe decisions. Examples include system simulation with guaranteed accuracy \cite{lauricella2020set} and controller tuning algorithms that avoid unreliable parameters \cite{berkenkamp2016safe,lederer2020parameter}. By establishing our optimal, non-asymptotic uncertainty bounds, our work aims at bridging non-parametric kernel learning and robust analysis and control. Practical applications of the results include predictive control schemes that explicitly incorporate non-parametric uncertainties \cite{manzano2020robust,maddalena2020kpc}, and the certification of machine learning-based algorithms \cite{knuth2021planning,wabersich2021probabilistic}, but in a deterministic fashion. More generally, the provided tools could also be employed in \red{the domain of real-time optimization under unknown constraints \cite{chachuat2009adaptation}}. Our motivation is similar in essence to the ones found in non-linear set membership and interval analysis works \cite{raissi2004set,karimshoushtari2020design}, but our study is focused on kernel machines and their associated spaces.

\textbf{Notation:} $\mathbb{N}$ denotes the set of natural numbers and $\mathbb{R}^d$ is the $d$-dimensional Euclidean space. Let $S_1$ and $S_2$ be subspaces of $S$, then $S_1 \oplus S_2 = S$ indicates their vector direct sum, i.e., $\forall s \in S$, $\exists! s_1 \in S_1, \exists! s_2 \in S_2 : s = s_1 + s_2$. We denote by $K_{XX}$ the matrix of kernel evaluations at $X$, i.e., the matrix containing $k(x_i,x_j)$ at its $i$-th row and $j$-th column, $x_i,x_j \in X$. Let a query point $x$ be specified, then $K_{Xx}$ represents the column vector-valued function $x \mapsto \begin{bmatrix} k(x,x_1) & \dots & k(x,x_d) \end{bmatrix}^\top \in \mathbb{R}^d$, whereas $K_{xX}$ denotes its transpose. For a matrix $A$ we denote its nullspace by $N(A)$.

\section{Preliminary: Kernels and their \red{RKHSs}}
\label{sec::Preliminary}
We start by briefly reviewing the formalism of kernel learning, and define the main elements that will be later used in our analysis. The reader is referred to \cite{schaback2006kernel,manton2015primer} for further details on \red{this} topic.

A kernel $k:\Omega\times\Omega\rightarrow \mathbb{R}$ is any symmetric, real-valued function defined on a non-empty input set $\Omega$. $k$ is said to be positive-definite if the weighted sum $\sum_{i=1}^n \sum_{j=1}^n \alpha_i \alpha_j k(x_i,x_j)$ is strictly positive $\forall n \in \mathbb{N}$, $\forall \alpha_1,\dots,\alpha_n \in \mathbb{R}\backslash \{0\}$, $\forall x_1,\dots,x_n \in \Omega$. An example of a commonly used continuous kernel that enjoys this property is the squared-exponential, also known as the radial basis function (RBF) kernel. Associated with each $k$, there is a unique Hilbert space of maps $\mathcal{H}$ that is referred to as the reproducing kernel Hilbert space (RKHS) of $k$. For compact domains $\Omega$, \cite{micchelli2006universal} presents families of kernels whose $\mathcal{H}$ are dense in the space of continuous functions, which can be interpreted as a measure of richness of such a space. Let $\mathbb{R}^\Omega = \{f:\Omega \rightarrow \mathbb{R} \}$ and $L_x:\mathbb{R}^\Omega \rightarrow \mathbb{R}$ be the map $L_x:f\mapsto f(x)$, also known as the evaluation functional for a given $x \in \Omega$. Formally, a RKHS is simply a Hilbert space $\mathcal{H} \subset \mathbb{R}^\Omega$ for which the $L_x$ maps are continuous $\forall x \in \Omega$. It turns out that partially-evaluated kernels $k(x,\cdot)$ belong to $\mathcal{H}$, $\forall x \in \Omega$ and define evaluation functionals through $\rkinner{f}{k(x,\cdot)} = f(x)$, $\forall f \in \mathcal{H}, \forall x \in \Omega$. The latter is known as the reproducing property. From a constructive viewpoint, $\mathcal{H}$ is given as the closure (w.r.t. the topology induced by the inner-product) of $\text{span}\left(\left\{ k(x,\cdot), x\in\Omega \right\} \right)$, encompassing thus weighted sums of partial kernels and limit points of sequences as well. It can be shown that this construction results in proper functions and not in equivalence classes of them.

Let $f \in \mathcal{H}$ with finite expansion $f = \sum_{i=1}^{n_f} \alpha_i k(x_i,\cdot)$, $\alpha_i \in \mathbb{R}$, $x_i \in \Omega$, for all $i$. Its induced norm $\rknorm{f}$ is then given by 
\begin{equation}
\begin{aligned}
    \rknorm{f}^2 =& \left\langle \sum_{i=1}^{n_f} \alpha_i k(x_i,\cdot),\sum_{i=1}^{n_f} \alpha_i k(x_i,\cdot) \right\rangle_\mathcal{H}\\ =& \sum_{i=1}^{n_f} \sum_{j=1}^{n_f} \alpha_i \alpha_j k(x_i,x_j) = \alpha^\top K_{XX} \alpha
\end{aligned}
\end{equation}
due to the reproducing property, with $\alpha$ being the vector of scalar weights. If a member $f \in \mathcal{H}$ is the limit of a sequence $f = \sum_{i=1}^\infty \alpha_i k(x_i,\cdot)$, then its norm is $$\rknorm{f}^2 = \lim_{n\rightarrow \infty} \sum_{i=1}^{n} \sum_{j=1}^{n} \alpha_i \alpha_j k(x_i,x_j).$$ As a last introductory step, we consider a finite subset $X \subset \Omega$ and define the power function $P_X:\Omega \rightarrow \mathbb{R}_{\geq0}$ as
\begin{equation}
	P_X(x) = \sqrt{k(x,x) - K_{xX}K_{XX}^{-1}K_{Xx}}
\end{equation}
whenever clear from the context, the reference to $X$ will be omitted. $P_X(x)$ can be interpreted as a form of statistical covariance, and evaluates to zero $\forall x \in X$.

\section{Optimal bounds in RKHSs}
\label{sec::OptimalBound}
\textcolor{black}{This section introduces the main technical results of this work. First, an infinite-dimensional variational problem is formulated to bound the ground-truth values at unseen locations, and its equivalence to a finite-dimensional problem is shown. Then, we discuss properties of the derived bounds and a closed-form alternative that does not involve solving any optimization problem.}

Herein we consider a positive-definite kernel $k:\Omega \times \Omega \rightarrow \mathds{R}$ along with its corresponding RKHS $\mathcal{H} \subset \mathds{R}^\Omega$. Our input space is taken to be a compact subset of the Euclidean space $\Omega \subset \mathbb{R}^n$. A dataset $\{(x_i,\mathsf{y}_i)\}_{i=1}^d$ is given to us, being composed of inputs $x_i \in \Omega$ and outputs $\mathsf{y}_i \in \mathbb{R}^{n_i}$, $\mathsf{y}_i = \begin{bmatrix}y_{i,1} & \dots & y_{i,n_i} \end{bmatrix}^\top$ that contain $n_i$ scalar samples collected at the same input location $x_i$. The dataset carries information about an underlying ground-truth map $f^\star \in \mathcal{H}$ according to
\begin{equation}
	y_{i,j} = f^\star(x_i) + \delta_{i,j}
\end{equation}
where $\delta_{i,j}$ represents an additive measurement noise that is assumed to be uniformly bounded \textcolor{black}{as stated next.}
\begin{myAssumption}
\label{as:noisebound}
    The magnitude of each noise realization $\delta_{i,j}$ is bounded by a known scalar quantity $\bar{\delta}$, i.e. $|\delta_{i,j}| \leq \bar \delta, \forall i,j$.
\end{myAssumption}

\begin{myAssumption}
	An estimate $\Gamma \geq \rknorm{f^\star}$ for the ground-truth norm is known.
\end{myAssumption}

For notational convenience, we define the quantities $X := \{ x_1, \dots, x_d \}$ and $\mathsf{y} := \begin{bmatrix} \mathsf{y}_1^\top & \dots & \mathsf{y}_d^\top \end{bmatrix}^\top$, which represent respectively the collection of inputs and the available outputs.

The aim is to quantify the uncertainty associated with \red{values of} the latent function $f^\star$ in the output space $\mathbb{R}$. We note that, from the reproducing property and the Cauchy–Schwarz inequality, one can readily establish
\begin{subequations}
\begin{align}
|f(x)| &= |\rkinner{f}{k(x,\cdot)}| \\
&\leq \rknorm{f} \rknorm{k(x,\cdot)} \\
&= \rknorm{f} \, \sqrt{k(x,x)} < \infty  
\end{align}
\label{eq.uniformBound}
\end{subequations}
$\forall f \in \mathcal{H}$, including the ground-truth $f^\star$. If the kernel $k$ is translation-invariant, \red{then} $k(x,x)$ is constant $\forall x \in \Omega$ and \eqref{eq.uniformBound} constitutes a uniform trivial bound for the unknown function. A reason for this inequality to be rather loose is that it does not incorporate any information provided by the \red{outputs $\mathsf{y}$} or by the \red{quantity} $\bar \delta$, which we exploit next.

\textcolor{black}{To upper bound the ground-truth values, we consider} the following infinite-dimensional variational problem $\mathds{P}0$, with the query point $x \in \Omega$ as a parameter
\begin{equation}
	\text{F}(x) = \sup_{f \, \in \mathcal{H}} \, \{ f(x) : \rknorm{f} \leq \Gamma, \infnorm{f_X - \mathsf{y}} \leq \bar \delta \} 
	\label{eq.infdimprob}
\end{equation}
where $f_X := \Lambda \begin{bmatrix} f(x_1) & \dots & f(x_d) \end{bmatrix}^\top$ is the vector of evaluations at the input locations, which are repeated whenever multiple outputs are available at a given input. This is accomplished through $\Lambda$ as defined in Appendix~\ref{app.thebigmatrix}. We highlight that the supremum is guaranteed to exist thanks to \eqref{eq.uniformBound}. Given a query location $x$, $\mathds{P}0$ yields the tightest
upper bound for $f(x)$ over all members $f \in \mathcal{H}$ of our hypothesis space that are consistent with our dataset, as well as our knowledge on the ground-truth complexity $\rknorm{f} \leq \Gamma$. Note how linking the function evaluations $f_X$ and the outputs $\mathsf{y}$ plays a role analogous to conditioning stochastic processes on past observations in statistical frameworks.

Consider now the convex parametric quadratically-constrained linear program $\mathds{P}1$ 
\begin{subequations}
	\label{eq.P2case1}
	\begin{align}
		{\normalfont \text{C}(x)} \; = \max_{c \in \mathds{R}^d, c_x \in \mathds{R}} & \quad c_x \\ 
		\text{\normalfont subj. to} \; & \ \; 
		\begin{bmatrix}
			c \\
			c_x
		\end{bmatrix}^\top 
		\begin{bmatrix}
			K_{XX} & K_{Xx} \\
			K_{xX} & k(x,x)
		\end{bmatrix}^{-1} 
		\begin{bmatrix}
			c \\
			c_x
		\end{bmatrix} \leq \Gamma^2 \label{eq.P2constrA} \\
		& \ \; \; \infnorm{\Lambda c - \mathsf{y}} \leq \bar\delta \label{eq.P2constrB}
	\end{align}
	\label{eq.P2}%
\end{subequations}
for any $x \in \Omega \backslash \{X\}$, and extend its value function to points \red{from the dataset} $x = x_i \in X$ with the solution of $\mathds{P}1': \, \text{C}(\red{x_i}) \,=\, \max_{c \in \mathbb{R}^d} \{c_i \, | \, c^\top K_{XX}^{-1}c \leq \Gamma^2, \, \infnorm{\Lambda \, c - \mathsf{y}} \leq \bar \delta \}$, \red{where $c_i$ is the $i$-th component of $c$.}
\textcolor{black}{This can be thought of as finding a map that interpolates the points $\{(x_i, c_i)\}_{i=1}^d$ and maximizes its value $c_x$ at the input location $x$.} The two cases \textcolor{black}{$\mathds{P}1$ and $\mathds{P}1'$} are distinguished \textcolor{black}{due to the matrix in \eqref{eq.P2constrA} becoming singular for any $x\in X$, and }since it allows for one decision variable to be eliminated. \red{Finally}, the connection between \eqref{eq.infdimprob} and \eqref{eq.P2} is stated next.

\begin{theorem} {\normalfont \textbf{(Finite-dimensional equivalence):}}
	\label{thm.main}
	The objective in $\mathds{P}0$ attains its supremum in $\mathcal{H}$ and ${\normalfont \text{F}}(x) = {\normalfont \text{C}}(x)$ for any $x \in \Omega$.
\end{theorem}

The derivation of the result, \red{which is given in Appendix~\ref{app.thmproof}}, follows lines similar to classical representer theorem ones, i.e., showing that the optimizer necessarily lies in a finite-dimensional subspace of the RKHS. Nevertheless, note that the objective $\mathds{P}1$ is not regularized, nor is $x$ necessarily an input of our dataset. Moreover, the proof also establishes the attainment property in $\mathcal{H}$, which helps in understanding the nature of the constraints.

Complementing \eqref{eq.infdimprob}, one could also be interested in the infimum $\inf_{f \, \in \mathcal{H}} \, \{ f(x) : \rknorm{f} \leq \Gamma, \infnorm{f_X - \mathsf{y}} \leq \bar \delta \}$ bounding the lowest attainable value at $x$. In this case, a result analogous to Theorem~\ref{thm.main} could be established, showing its equivalence to 
\begin{subequations}
	\begin{align}
		{\normalfont \text{B}(x)} \; = \min_{c \in \mathds{R}^d, c_x \in \mathds{R}} & \quad c_x  \\ 
		\text{\normalfont subj. to} \; & \ \; 
		\begin{bmatrix}
			c \\
			c_x
		\end{bmatrix}^\top 
		\begin{bmatrix}
			K_{XX} & K_{Xx} \\
			K_{xX} & k(x,x)
		\end{bmatrix}^{-1} 
		\begin{bmatrix}
			c \\
			c_x
		\end{bmatrix} \leq \Gamma^2 \label{eq.P3constrA} \\
		& \ \; \; \infnorm{\Lambda \, c - \mathsf{y}} \leq \bar\delta \label{eq.P3constrB}
	\end{align}
	\label{eq.BofX}%
\end{subequations}
for any $x\in\Omega \backslash X$, and extended to $\text{B}(\red{x_i}) = \min_{c \in \mathbb{R}^d} \{c_i \, | \, c^\top K_{XX}^{-1}c \leq \Gamma^2, \, \infnorm{\Lambda \, c - \mathsf{y}} \leq \bar \delta \}$ for $x=x_i \in X$. As a result of this subsection, for any point in the domain $x\in\Omega$, the solutions to the two convex programs \eqref{eq.P2} and \eqref{eq.BofX} define an \textit{uncertainty envelope} that confines the ground-truth to its interior $B(x) \leq f^\star(x) \leq C(x)$. 

\begin{myRemark}
	{\normalfont \textbf{(On the necessity of $\Gamma$):}} Data alone are not sufficient to compute \textit{any} out-of-sample bounds when considering functions $f \in \mathcal{H}$, regardless of the number of samples $d < \infty$ that one has. Given any tentative bound $\epsilon$ at $x \not\in X$, there exists $f_\rho \in \mathcal{H}$ consistent with the dataset that will violate the bound, that is, $f_\rho(x) > \epsilon + \rho$, for any pre-specified violation level $\rho > 0$. This is simply due to the existence of maps that can interpolate any finite set of samples. Restricting the search to the $\Gamma$-ball in $\mathcal{H}$ limits the flexibility of the considered functions, thus allowing for guarantees to be established. An analogous argument can be made in the space of Lipschitz functions. If no bound is posed on the Lipschitz constant of the ground-truth, assuming Lipschitz continuity \textit{per se} becomes vacuous. 
\end{myRemark}

\textcolor{black}{
\begin{myRemark}\normalfont \textbf{(On the noise assumption):} Assumption~\ref{as:noisebound} is central to the robust control literature and requires a careful handling in practice. Quantities $\bar \delta$ estimated from historical data could be invalidated by newly obtained samples, potentially rendering $\mathds{P}1$ infeasible. In such a case, $\bar{\delta}$ would need to be augmented to accommodate the new samples (similar issues are discussed in \cite[Section~3]{milanese2004set}). Certainly, dataset pre-processing and outlier detection are essential to the success of data-driven methods, including the present one, in practice.
\end{myRemark}}

\textcolor{black}{
\begin{myRemark}\normalfont \textbf{(On having loose $\Gamma$ and $\bar \delta$):} As formulated in \eqref{eq.infdimprob}, the bound $\text{F}(x)$ depends on the quality of the available $\Gamma$ and $\bar \delta$. The looser these two quantities are, the larger the resulting bounds|see a numerical example in Section~\ref{sec::Result}. A straightforward method is presented in Appendix~\ref{app.normEst} to compute RKHS norm lower estimates purely based on data, which are refined the more samples one has. The methodology can help users to arrive at upper bounds $\Gamma$ through augmentation.
\end{myRemark}}

\subsection{Width and width shrinkage}

\begin{figure}[b!]
	\centering
	\includegraphics[scale=0.38]{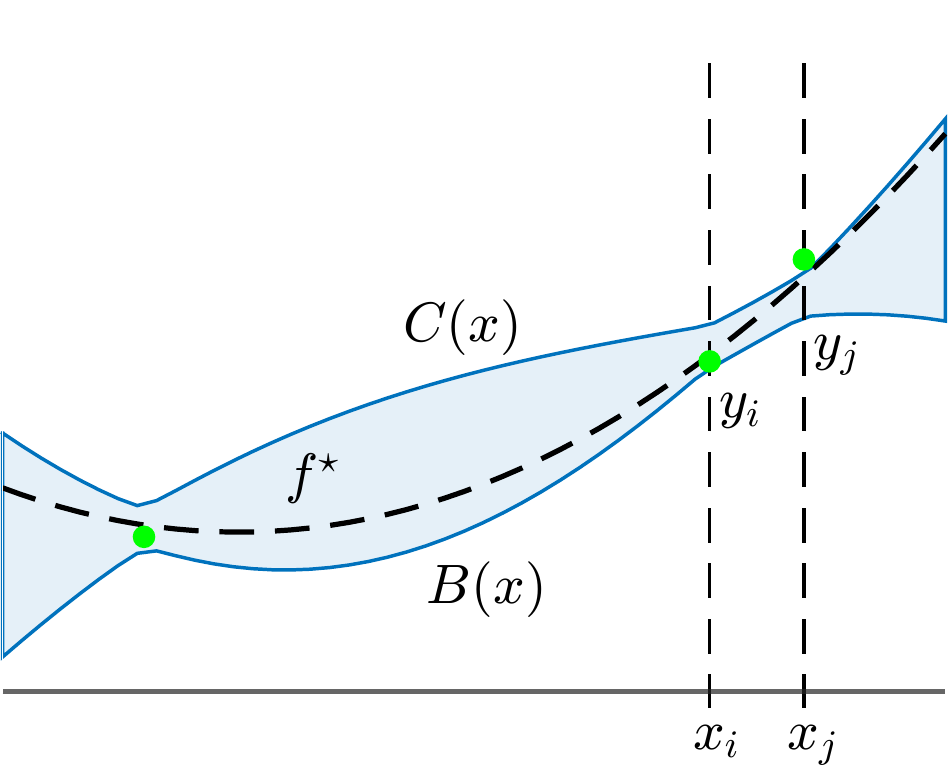} \hspace{8pt}
	\includegraphics[scale=0.38]{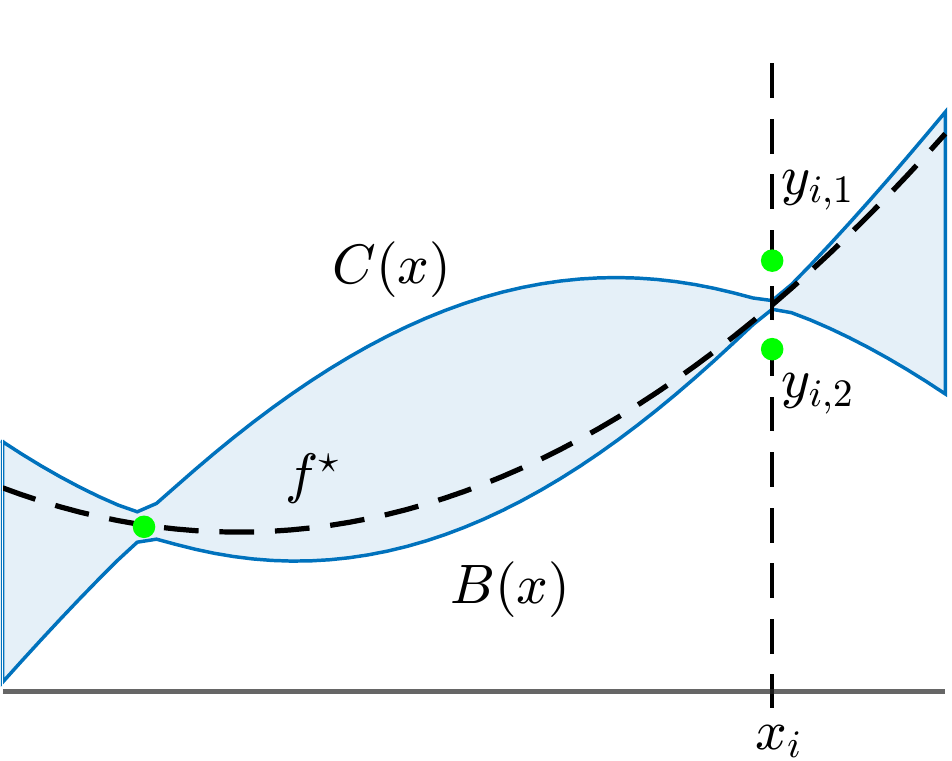}
	\caption{(Left) A sample lying outside of the uncertainty envelope, implying that the width is smaller than $\bar \delta$ at $x_j$. (Right) Redundant information is used to shrink the uncertainty envelope. In this scenario, we recover the ground-truth value at $x_i$ as $\text{C}(x_i) = \text{B}(x_i) = f^\star(x_i)$.}
	\label{fig.riskBound}
\end{figure}

Given our knowledge on the noise influence $\bar \delta$, it is natural to ask what the limits of the uncertainty quantification technique considered herein are. For example, is the width of the envelope $\text{C}(x) - \text{B}(x)$ restricted to a certain minimum value that cannot be reduced even with the addition of new data? From \eqref{eq.P2constrB}, it is clear that at any input location $x_i \in X$, $\text{C}(x_i)$ and $\text{B}(x_i)$ cannot be more than $2\bar \delta$ apart. In addition to that, the presence of the complexity constraint \eqref{eq.P2constrA} can bring the two values closer to each other. Depending on how restrictive this latter constraint is for a given $x_i$, the corresponding output $y_i$ might lie outside the interval between $\text{C}(x_i)$ and $\text{B}(x_i)$. In this case, the resulting width is considerably reduced as illustrated in Figure~\ref{fig.riskBound} (left).

\begin{proposition} 
	\label{prop.smallWidth}
	{\normalfont \textbf{(Width smaller than the noise bound):}}
	If $\exists \mathsf{y}_i$ such that $y_{i,j} > {\normalfont C}(x_i)$ or $y_{i,j} < \text{B}(x_i)$ for some $j$, then $\text{C}(x_i) - \text{B}(x_i) \leq \bar \delta$.
\end{proposition}
\begin{myProof}
	Follows from $\text{C}(x_i) \geq \text{B}(x_i)$, $\text{C}(x_i) \leq y_{i,j} + \bar \delta$ and $\text{B}(x_i) \geq y_{i,j} - \bar \delta$ for any $i=1,\dots,d$ and any $j=1,\dots,n_i$.\QED 
\end{myProof} 
Suppose now one has sampled $(x_i,\mathsf{y}_i)$ with $\mathsf{y}_i = \begin{bmatrix} y_{i,1} & y_{i,2} \end{bmatrix}^\top$, $y_{i,1} = f^\star(x_i) + \bar \delta$ and $y_{i,2} = f^\star(x_i) - \bar \delta$. Then there is no uncertainty whatsoever about $f^\star$ at $x_i$ since $f^\star(x_i) = (y_{i,1} + y_{i,2})/2$ is the only possible value attainable by the ground-truth. This illustrates that the possibility of having multiple outputs at the same location allows for the uncertainty interval to shrink past the $\bar \delta$ width, and eventually even reduce to a singleton as shown in Figure~\ref{fig.riskBound} (right). Notwithstanding, the addition of a new datum to an existing dataset|be it in the form of a new output at an already sampled location or a completely new input-output pair|can only reduce the uncertainty.

\begin{proposition} {\normalfont \textbf{(Decreasing uncertainty):}}
	Let $C_{{\normalfont 1}}(x)$ denote the solution of $\mathds{P}1$ with a dataset $D_1 = \{(x_i,\mathsf{y}_i)\}_{i=1}^{d}$, and $C_{{\normalfont 2}}(x)$ the solution with $D_2 = D_1 \cup \{(x_{d+1},\mathsf{y}_{d+1})\}$. Then $C_{{\normalfont 2}}(x) \leq C_{{\normalfont 1}}(x)$ for any $x \in \Omega$. 
	\label{prop.decreasing}
\end{proposition}

\begin{myProof}
	Denote by $\mathds{P}1_1$ the problem solved with $D_1$ and decision variables $\begin{bmatrix} c & c_x \end{bmatrix}$. Similarly, $\mathds{P}1_2$ is associated with the dataset $D_2$ and the decision variables $\begin{bmatrix} c & c_x & c_z \end{bmatrix}$, where $c_z$ are due to the additional input in $D_2$. Since $D_2$ contains all members of $D_1$, the $\infty$-norm constraint of $\mathds{P}1_2$ can be recast as that of $\mathds{P}1_1$ and an additional constraint for $c_z$ and the new outputs. Let $\red{\mathds{X}}  := X \cup \{x\}$, $\bar c := \begin{bmatrix} c^\top c_x \end{bmatrix}^\top$ and $z := x_{d+1}$ be shorthand variables to ease notation. The complexity constraint of $\mathds{P}1_2$ is then
	\begin{subequations}
		\begin{align}
			&\;\;\begin{bmatrix}
				\bar c \\
				c_z
			\end{bmatrix}^\top 
			\begin{bmatrix}
				K_{\red{\mathds{X}} \red{\mathds{X}}} & K_{\red{\mathds{X}} z} \\
				K_{z \red{\mathds{X}}} & k(z,z)
			\end{bmatrix}^{-1} 
			\begin{bmatrix}
				\bar c \\
				c_z
			\end{bmatrix} \leq \Gamma^2 \\
			\overset{(i)}{\Leftrightarrow}& \;\;
			\bar c^\top K_{\red{\mathds{X}} \red{\mathds{X}}}^{-1} \bar c +
			P_{\red{\mathds{X}}}^{-2}(z) \, 
			\left\|
			\begin{bmatrix}
				K_{\red{\mathds{X}} \red{\mathds{X}}}^{-1} K_{\red{\mathds{X}} z} \\
				-1
			\end{bmatrix}
			\begin{bmatrix}
				\bar c \\
				c_z
			\end{bmatrix}\right\|_2^2
			\leq \Gamma^2 \\
			\overset{(ii)}{\Leftrightarrow}& \;
			\begin{bmatrix}
				c \\
				c_x
			\end{bmatrix}^\top 
			\begin{bmatrix}
				K_{XX} & K_{Xx} \\
				K_{xX} & k(x,x)
			\end{bmatrix}^{-1} 
			\begin{bmatrix}
				c \\
				c_x
			\end{bmatrix} \label{eq.tightenedConstr}\\\notag
			&\qquad\qquad\qquad+
			P_{\red{\mathds{X}}}^{-2}(z) \, \left( \bar c^\top K_{\red{\mathds{X}} \red{\mathds{X}}}^{-1} K_{\red{\mathds{X}} z} - c_z \right)^2
			\leq \Gamma^2 
		\end{align}
		\label{eq.quadraticDecomp}
	\end{subequations}
	where the matrix identity found in Appendix~\ref{app.blockMatrixIdent} was used in $(i)$ and $P^2_{\red{\mathds{X}}}(z) = k(z,z) - K_{z \red{\mathds{X}}} K_{\red{\mathds{X}} \red{\mathds{X}}}^{-1} K_{\red{\mathds{X}} z}$. In $(ii)$, the definitions of $\bar c$ and $\red{\mathds{X}}$ were used. Thanks to $P_{\red{\mathds{X}}}(z) \geq 0, \forall z$ and the quadratic term multiplying it, we conclude that for any choice of the decision variable $c_z$, \eqref{eq.tightenedConstr} is a tightened version of the complexity constraint of $\mathds{P}1_1$, which is \eqref{eq.P2constrA}. As a result, the maximum of $\mathds{P}1_2$ is lower or equal than that of $\mathds{P}1_1$. \QED 
\end{myProof}

Let us take a closer look at the tightened constraint \eqref{eq.tightenedConstr}. The term $\bar c^\top K_{\red{\mathds{X}} \red{\mathds{X}}}^{-1} K_{\red{\mathds{X}} z} =: s(z)$ represents an interpolating model passing through the output values $\bar c$, that is, $c$ and $c_x$ (see e.g. the discussion in Section~3.1 of \cite{maddalena2020deterministic}). If the difference $s(z) - c_z$ can be made small, then the tightening will also be reduced, whereas it will be significant if the difference is large. The result is of course dictated by the $\infty$-norm constraint, since $c_z$ cannot be more than $\bar \delta$ away from all the outputs $\mathsf{y}$ available at $z$. Therefore, a new datum will cause significant shrinkage of the envelope at a point $z \in \Omega$ when the new output causes $s(z) - c_z$ to be large\textcolor{black}{, which intuitively can be seen as a measure of gained information through the new sample.} Finally, this process is weighted by the inverse of the power function $P_{\red{\mathds{X}}}^{-2}(z)$, which does not depend on any output, but only on the input locations. \red{
For more practical guidelines and a visual representation of how new data can contribute to reducing the ground-truth uncertainty, the reader is referred to Examples~1 and 2 from Section~\ref{sec::Result}.
}

\begin{myRemark}
	Recovering the ground-truth as shown in Figure~\ref{fig.riskBound} (right) requires the noise realizations to match $\bar \delta$ and $-\bar \delta$; it is thus necessary to have tight noise bounds for it to happen. On the other hand, Proposition~\ref{prop.decreasing} guarantees the decreasing uncertainty property regardless of how \textcolor{black}{accurate} $\bar \delta$ is. Although not explicitly stated, a completely analogous result holds for the lower part of the envelope $\text{B}(x)$.
\end{myRemark}

\subsection{A sub-optimal closed-form alternative}

The discussion in this subsection assumes that only one sample is present at each input location, i.e., $\mathsf{y}_i = y_i$ for $i=1,\dots,d $, so that $\mathsf{y}=y$.

In order to alleviate the computational complexity of having to solve two optimization problems at each query point, closed-form expressions can be \red{employed instead}. These surrogates yield sub-optimal bounds around \red{any pre-specified kernel model of the form} $s(x) = \alpha^\top K_{Xx}$, \red{for some} $\alpha \in \mathbb{R}^d$. 
\begin{proposition}
	\label{prop:uniform}
	Let $s(x)=\alpha^\top K_{Xx}$, \red{for a given} $\alpha \in \mathbb{R}^d$. Then, for any $x\in \Omega$, the ground-truth is bounded by $s(x) - S(x) \leq f^\star(x) \leq s(x) + S(x)$ with
	\begin{equation}
		S(x) =  P_X(x) \, \sqrt{ \Gamma^2 + \tilde\Delta } + \bar{\delta} \, \onenorm{K_{XX}^{-1}K_{Xx}} + \, \vert \tilde{s}(x) - s(x) \vert
		\label{eq.uniformBound2}
	\end{equation}
	where $\tilde{s}(x) = y^\top K_{XX}^{-1} K_{Xx}$, and the constant $\tilde\Delta$ is the \red{minimum} of the unconstrained convex problem $\min_{\nu \in \mathbb{R}^d} \left\{ \frac{1}{4}\nu^\top K_{XX}\nu + \nu^\top y + \bar{\delta} \onenorm{\nu}\right\}$.
\end{proposition}
\begin{myProof}
	See Appendix~\ref{app.lemproof}.
\end{myProof}

The map $\tilde{s}(x) = y^\top K_{XX}^{-1} K_{Xx}$ is an interpolant for the available outputs $y$. Note also that none of the terms in \eqref{eq.uniformBound2} depend on the model weights $\alpha$ with the exception of the last term $|\tilde s(x) - s(x)|$. Therefore, the width $S(x)$ will be minimized when $s(x) = \tilde s(x) \implies \alpha = y^\top K_{XX}^{-1}$. Since such a model would severely overfit, a balance between smoothing the data and not diverging too much from $\tilde s(x)$ has to be found. In our previous work \cite{maddalena2020deterministic}, we have illustrated how kernel ridge regression and minimum norm models are good candidate techniques to accomplish this goal.

By reformulating the optimal bounds, we uncover their relation with the suboptimal estimates given in Proposition~\ref{prop:uniform}. First, consider $\mathds{P}1$ and optimize over the decision variable $\delta = c - y$ rather than over $c$. Next, apply a quadratic decomposition identical to the one used in \eqref{eq.quadraticDecomp} to the complexity constraint \eqref{eq.P2constrA} and solve for $c_x$. After recalling that $\tilde{s}(x) = y^\top K_{XX}^{-1} K_{Xx}$ and $\rknorm{\tilde s}^2 = y^\top K_{XX}^{-1} y$, one obtains
\begin{equation}
\begin{aligned}
c_x  \leq \;&  P(x) \sqrt{ \Gamma^2 - \rknorm{\tilde s}^2 - \delta^\top K_{XX}^{-1}\delta +2y^\top K_{XX}^{-1}\delta}\\
&+\tilde s(x) 
+ \delta^\top K_{XX}^{-1}K_{Xx}
\end{aligned}	
\label{eq.ineq}
\end{equation}
Instead of maximizing $c_x$, the right-hand side of \eqref{eq.ineq} can be directly considered as the objective function equivalently. As a result, we obtain 
\[
\begin{aligned}
\max_{\infnorm{\delta} \leq \bar \delta}\;\;& \tilde s(x) + P(x) \sqrt{ \Gamma^2 - \rknorm{\tilde s} - \delta^\top K_{XX}^{-1}\delta -2y^\top K_{XX}^{-1}\delta}\\
&+ \delta^\top K_{XX}^{-1}K_{Xx} 
\end{aligned}
\]
Now, relax the problem by allowing $\delta$ to attain different values inside and outside the square-root
\begin{subequations}
	\begin{align}\notag
		\max_{\delta_1,\delta_2 \in \mathds{R}^{d}} &  \tilde s(x) + P(x) \sqrt{ \Gamma^2 - \rknorm{\tilde s}^2 - \delta_1^\top K_{XX}^{-1}\delta_1 +2y^\top K_{XX}^{-1}\delta_1}\\\label{eq.relaxedObjective}
		&\;\; + \delta_2^\top K_{XX}^{-1}K_{Xx} \\
		\text{\normalfont subj. to} & \;\; \; \infnorm{\delta_1} \leq \bar \delta, \; \infnorm{\delta_2} \leq \bar \delta 
	\end{align}
\end{subequations}
Note that the objective is separable and that $\tilde\Delta$ is the dual solution of 
$\max_{\delta_1 \in \mathbb{R}^d} \left\{ -\delta_1^\top K_{XX}^{-1}\delta_1 + 2y^\top K_{XX}^{-1}\delta_1 - \rknorm{\tilde s}^2 \right\}$.
Also, $\max_{\delta_2 \in \mathbb{R}^d} \{ \delta_2^\top K_{XX}^{-1}K_{Xx} : \infnorm{\delta_2} \leq \bar\delta \} = \bar{\delta} \onenorm{K_{XX}^{-1}K_{Xx}}$ since these norms are duals of each other. Remember that the objective \eqref{eq.relaxedObjective} is a conservative upper bound for $f^\star(x)$, having $\tilde s(x)$ as the reference model. Given any smoother $s(x)$, the triangle inequality $|f(x) - s(x)| \leq |f(x) - \tilde s(x)| + |\tilde s(x) - s(x)|$ can be used to bound the distance between its predictions and the ground-truth values, arriving thus at the same expressions presented in Proposition~\ref{prop:uniform}.

From \eqref{eq.ineq}, the noise variable $\delta$ is seen to increase the maximum in two distinct ways: through the inner product $\delta^\top K_{XX}^{-1}K_{Xx}$, and via a norm augmentation corresponding to $\tilde\Delta$. One source of conservativeness in Proposition~\ref{prop:uniform} is taking into account the worst-possible inner-product and norm increase jointly. Despite this fact, they yield competitive results for moderate noise-levels \red{as shown numerically in Section~\ref{sec::Result}}. We moreover note that in the noise-free scenario, \eqref{eq.ineq} and \eqref{eq.relaxedObjective} are the same, and Proposition~\ref{prop:uniform} simplifies to the classical bounds in the interpolation case (see for instance \cite{fasshauer2011positive}).

\begin{myRemark}
	The \red{sub-optimal} bounds presented in this subsection feature a nominal model at their center, which is desirable in many practical situations. In the optimal scenario, the minimum norm regressor $s^\star(x) = \alpha^{\star\top} K_{Xx}$, $\alpha^\star = \argmin_{\alpha \in \mathbb{R}^d} \{ \alpha^\top K_{XX} \alpha : \infnorm{K_{XX} \alpha - y} \leq \bar\delta \}$ can be used as a nominal model. This choice is guaranteed to lie completely within $\text{C}(x)$ and $\text{B}(x)$|although not necessarily in the middle|since the map $s^\star$ belongs to $\mathcal{H}$ and is a feasible solution for $\mathds{P}0$.
\end{myRemark}

\section{Efficient computation and outer approximations}
\label{sec::Approximation}
One of the fundamental sources of computational complexity in kernel learning lies in the inverse term $K_{XX}^{-1}$. Scaling these techniques to large datasets in a principled manner is still a topic of active research \cite{zhang2013divide,burt2020convergence}. Notice that $K_{XX}^{-1}$ is explicitly present in $\mathds{P}1$, thus limiting its applicability to small and medium-sized problems due to the cubic time complexity associated with the inverse operation. In this section we discuss alternative formulations that can be solved more efficiently.

\subsection{The dual problem}
\label{sec:dual}
Following a standard dualization procedure, which can be found in Appendix~\ref{app.dual}, a Lagrangian dual for $\mathds{P}1$ in \eqref{eq.P2} can be the convex problem $\mathds{D}1$ 
\begin{equation}
\begin{aligned}
\min_{\nu \in \mathbb{R}^{\tilde d}, \, \lambda > 0} \;\;& \frac{1}{4\lambda} \nu^\top \Lambda K_{XX} \Lambda^\top \nu + \left(\mathsf{y} - \frac{1}{2\lambda} \Lambda K_{Xx} \right)^\top \nu\\
&+ \bar \delta \Vert \nu \Vert_1 +  \frac{1}{4\lambda} k(x,x) + \lambda \Gamma^2    
\end{aligned}
\label{eq.dualProb}
\end{equation}
for any query point $x \in \Omega \backslash X$. In our notation the dimension $\tilde d = \sum_{i=1}^d n_i$ is the total number of outputs, that is, the size of $\mathsf{y}$. As detailed in Appendix~\ref{app.dual}, under the assumption of the complexity constrain\textcolor{black}{t} not being active, the dual of $\mathds{P}1'$ is also $\mathds{D}1$, meaning that the formulation \eqref{eq.dualProb} could be used $\forall x \in \Omega$.

Remarkably, $\mathds{D}1$ only involves the kernel matrix itself and not its inverse, avoiding thus the aforementioned adversity. Furthermore, the query point $x$ enters $\mathds{D}1$ through the terms $K_{Xx}$ and $k(x,x)$. The former measures the similarity between the query point $x$ and each of the inputs in $X$; the latter is simply a constant term for translation-invariant kernels, and evaluates always to 1 in the specific case of the squared-exponential kernel.  

The optimization problem above is convex since it is a quadratic-over-linear function with $\Lambda K_{XX} \Lambda^\top \succeq 0$ and $\lambda$ restricted to the positive reals. The objective can moreover be decomposed into a differentiable part and a single non-differentiable term $\onenorm{\nu}$, with $\nu$ unconstrained. This class of problems has long been studied and mature numerical algorithms exist to solve them, notably different flavors of splitting methods such as the alternating direction method of multipliers (ADMM) \cite[Section~6]{boyd2011distributed}. Alternatively, a standard linear reformulation could be employed to substitute $\onenorm{\nu}$ by $\sum_i \eta_i$, with additional constraints $-\nu \leq \eta, \, \nu \leq \eta$. The result is a completely differentiable objective, but with extra decision variables and linear constraints. Next, a mild condition is given ensuring a zero duality gap between the primal and dual problems.

\begin{proposition}
	{\normalfont \textbf{(Strong duality):}}
	If $\bar \delta > \delta_{i,j}, \forall i, j$ and $\Gamma > \rknorm{f^\star}$, then no duality gap exists, i.e., $\max \mathds{P}1 = \min \mathds{D}1$.
\end{proposition}

\begin{myProof}
	Consider the primal problem $\mathds{P}1$ and select $c = f^\star_X$ and $c_x = f^\star(x)$. Let $\mathds{X} := X \cup \{x\}$ and \textcolor{black}{$K_{\mathds{X}\mathds{X}}$} denote the kernel matrix associated with $\mathds{X}$. Thanks to the optimal recovery property \cite[Theorem 13.2]{wendland2004scattered}, $\begin{bmatrix} c^\top & c_x \end{bmatrix} \textcolor{black}{K_{\mathds{X}\mathds{X}}} \begin{bmatrix} c^\top & c_x \end{bmatrix}^\top$ $\leq \rknorm{f^\star}^2$, which in turn is strictly smaller than $\Gamma^2$ by assumption. Also, $\infnorm{\Lambda c - \mathsf{y}} = \infnorm{\Lambda f^\star_X - \mathsf{y}} = \infnorm{\begin{bmatrix} \delta_{1,1} & \dots & \delta_{2,1} & \dots \end{bmatrix}^\top} < \bar \delta$. Therefore, the ground-truth values constitute a feasible solution that lies in the interior of the primal problem feasible set. As a result, Slater's condition is met and, since the primal is convex, there is no duality gap.\QED 
\end{myProof}

\subsection{An alternating optimization procedure}
\label{sec.alternating}

Solving the dual problem to any accuracy leads to an over bound on $\text{C}(x)$ thanks to duality. In other words, any feasible sub-optimal solution of $\mathds{D}1$ establishes a conservative uncertainty estimate. This motivates the study of light methods that could trade-off computational time and accuracy. In what follows we describe a block coordinate minimization scheme to tackle the problem, which is later shown to yield reasonable results after only a small number of iterations.

Whenever $\lambda$ is fixed to a particular positive value $\lambda^* > 0$, the problem \eqref{eq.dualProb} simplifies to an unconstrained quadratic program (QP) in $\nu$ of the form $\min_{\nu\in\mathbb{R}^{\tilde{d}}}\tilde{C}_x(\lambda^*, \nu)$. On the other hand, if $\nu$ is fixed to $\nu^* \in \mathbb{R}^{\tilde d}$, the dual objective takes the form
\begin{equation}
	\label{eq:fixnu}
	\min_{\lambda \in \mathbb{R}_{> 0}} \tilde{C}_x(\lambda, \nu^*)= \min_{\lambda \in \mathbb{R}_{> 0}} \frac{c_1}{\lambda} + c_2 \lambda +c_3
\end{equation}
with the constants
\begin{equation}
\begin{aligned}
    c_1 = &\frac{1}{4}\begin{bmatrix}
		\Lambda^\top \nu^* \\ -1
	\end{bmatrix}^\top
	\textcolor{black}{K_{\mathds{X}\mathds{X}}}
	\begin{bmatrix}
		\Lambda^\top \nu^* \\ -1
	\end{bmatrix}\;,\\
	c_2 =& \Gamma^2\;,\;\;c_3 =\mathsf{y}^\top \nu^*+\bar\delta \onenorm{\nu^*}\;
\end{aligned}
\end{equation}
and $\textcolor{black}{K_{\mathds{X}\mathds{X}}} \succeq 0$. We have $\frac{\partial\tilde{C}_x(\lambda, \nu^*)}{\partial\lambda} = \frac{-c_1}{\lambda^2}+c_2$ which gives the candidate solution $\lambda^* = \sqrt{c_1/c_2}$ for \eqref{eq:fixnu}. We have $c_2>0$ and $c_1\ge 0$ for any $x\in \Omega$, and $c_1>0$ for any $x\in\Omega\setminus X$. Furthermore, $\lambda^*$ is indeed a minimizer of \eqref{eq:fixnu} for $x\in\Omega\setminus X$ since its curvature is positive, i.e., $\frac{\partial^2\tilde{C}_x(\lambda^*, \nu^*)}{\partial\lambda^2} > 0$. In closed-form, $\lambda^*$ takes the following form.
\begin{equation}
	\lambda^* = \frac{\sqrt{{\nu^*}^\top \Lambda K_{XX} \Lambda^\top \nu^*-2 (\Lambda K_{Xx})^\top \nu^* + k(x,x)}}{2\Gamma}
\end{equation}
Note that $\lambda^*=0$ is only possible if $\begin{bmatrix}
	\Lambda^\top \nu^* \\ -1
\end{bmatrix}$ is in the nullspace of matrix \textcolor{black}{$K_{\mathds{X}\mathds{X}}$}, which is only possible if $x\in X$. In this case, after fixing $\lambda^*=0$, the problem to solve for $\nu$ reduces to 
$$\min_{\nu}\;\; \mathsf{y}^\top \nu + \bar\delta \onenorm{\nu} \; \; \text{s.t.}\;\; \begin{bmatrix}
	\Lambda^\top\nu \\ -1
\end{bmatrix} \in \mathrm{Null}(\textcolor{black}{K_{\mathds{X}\mathds{X}}})\,.$$
We formulate the alternating optimization algorithm for a maximum number of iterations $L$ and a termination threshold $\epsilon$ in the following way.

\begin{algorithm}[htbp!]
	\label{alg::altMin}
	\SetAlgoLined
	\KwResult{Upper bound $\tilde{C}(x)$ of the ground-truth at point $x$}
	\textbf{Input}: $x$, $\lambda_0$, $L$, $\epsilon$\\
	$\lambda_0^* = \lambda_0$\\
	$k=0$\\
	\Do{$k<L$ \text{and} $\vert \lambda^*_k - \lambda^*_{k-1} \vert > \epsilon$}{
		$\nu^*$ = $\argmin_{\nu \in \mathbb{R}^{\tilde{d}}} \tilde{C}_x(\lambda_k^*,\nu) $\\
		$\lambda_{k+1}^* = \frac{\sqrt{{\nu^*}^\top \Lambda K_{XX} \Lambda^\top \nu^*-2 (\Lambda K_{Xx})^\top \nu^* + k(x,x)}}{2\Gamma}$\\
		$k = k+1$
	}
	$\tilde{C}(x) = \tilde{C}_x(\lambda_k^*, \nu^*)$
	\caption{Alternating minimization}
\end{algorithm}

\begin{myRemark}[Numerical properties]
	Recall the convex dual objective function \eqref{eq.dualProb}. Since the non-differentiable term $\onenorm{\nu}$ is separable and the remainder of the objective is differentiable, a tuple $(\nu^*,\lambda^*)$ that simultaneously minimizes both sub-problems also necessarily minimizes the whole objective \eqref{eq.dualProb}. For non-asymptotic sublinear convergence rates of alternating minimization algorithms applied to convex programs, the reader is referred to the work \cite{beck2015convergence}.
\end{myRemark}

\section{On the connections with Gaussian processes}
\label{sec::GP}
Before proceeding to a \red{first} numerical example, we contrast our uncertainty quantification technique with Gaussian processes (GPs). By putting them into perspective, we hope to improve the understanding of the theory developed herein.

In the Bayesian framework of GPs, kernels are used to \red{parametrize} the covariance between random variables in the input space $\Omega$ \cite{GPforML}. Deeper links also exist between the Hilbert space associated with such stochastic processes and the RKHS $\mathcal{H}$ corresponding to their kernels \cite{parzen1961approach}, in that there exists an isometric isomorphism connecting both spaces. Moreover, it is known that even though the mean function of a GP does belong to $\mathcal{H}$, its sample paths almost surely do not if $\text{dim}(\mathcal{H})=\infty$. Nonetheless, the same paths can belong to another RKHS|not necessarily the one associated with their kernel|with probability one (see \cite{kanagawa2018gaussian} for a comprehensive discussion on the topic). The latter phenomenon is known as Driscoll’s zero-one law. Although \red{they unveil fundamental properties of the GP framework}, deriving practical guidelines from these results requires care as intuition might lead to wrong conclusions when examining infinite-dimensional spaces.

The variance of a Gaussian process is a form of uncertainty quantification against outputs $y$ drawn from \red{the distribution conditioned on a query input $x$}.
\red{Some other GP works are more aligned with our setting and consider a ground-truth map, either assumed to be a GP sample path \cite{lederer2019uniform} or to belong to the corresponding kernel RKHS \cite{berkenkamp2017safe,fiedler2021practical}. The latter works bound the difference between the GP mean and the ground-truth values, making use of the GP standard deviation times a uniform scaling term.} \red{As for noise models, or marginal likelihood in statistical terms, the most widely adopted forms are the Gaussian or sub-Gaussian formats. The primary motivation behind this choice is analytical tractability since the GP variance does not admit a closed-form expression and has to be numerically approximated} in case other noise models are used (e.g. heteroscedastic Gaussian, log-Gaussian, Bernoulli) \cite[Chapter~9]{GPforML}.

Whether one should opt for the theory developed herein or for Gaussian processes-based techniques is truly a question of model selection.
\textcolor{black}{If used to analyze kernel models of dynamical systems, our approach would allow for the use of robust analysis and control tools since worst-case effects and distances can be computed. On the other hand, carrying out modeling through Gaussian processes requires users to use stochastic control theory \cite{hewing2019cautious}.} As a result, the final yield should also be taken into account when choosing a technique. Do probabilistic inequalities suffice or does my application require deterministic certification? In our view, the deterministic and the stochastic frameworks have their own merits and the user should judge which of them is more adequate to tackle the problem at hand. \red{Despite the differences between the standing assumptions of the methods, we provide the reader with a comparison between the bounds developed herein and a popular GP alternative in Section~\ref{sec::Result}}.

\begin{figure*}[t]
	\centering
	\includegraphics[scale=0.38]{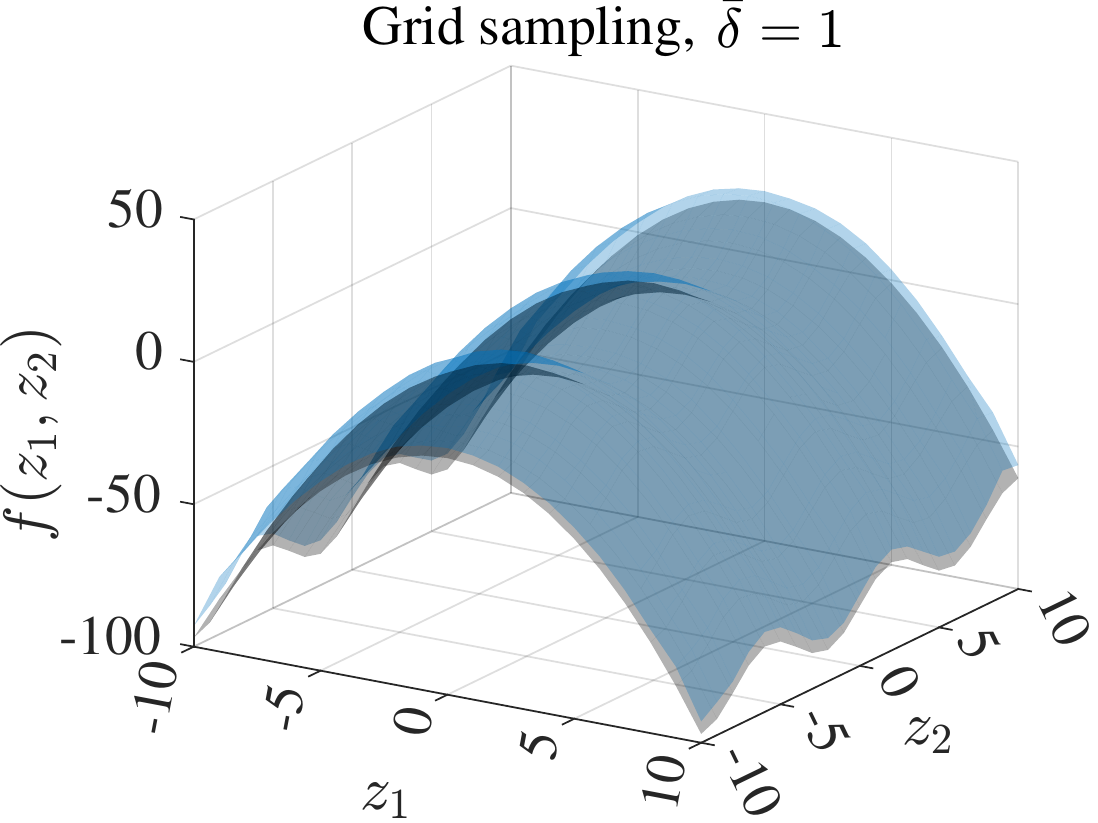}
	\includegraphics[scale=0.38]{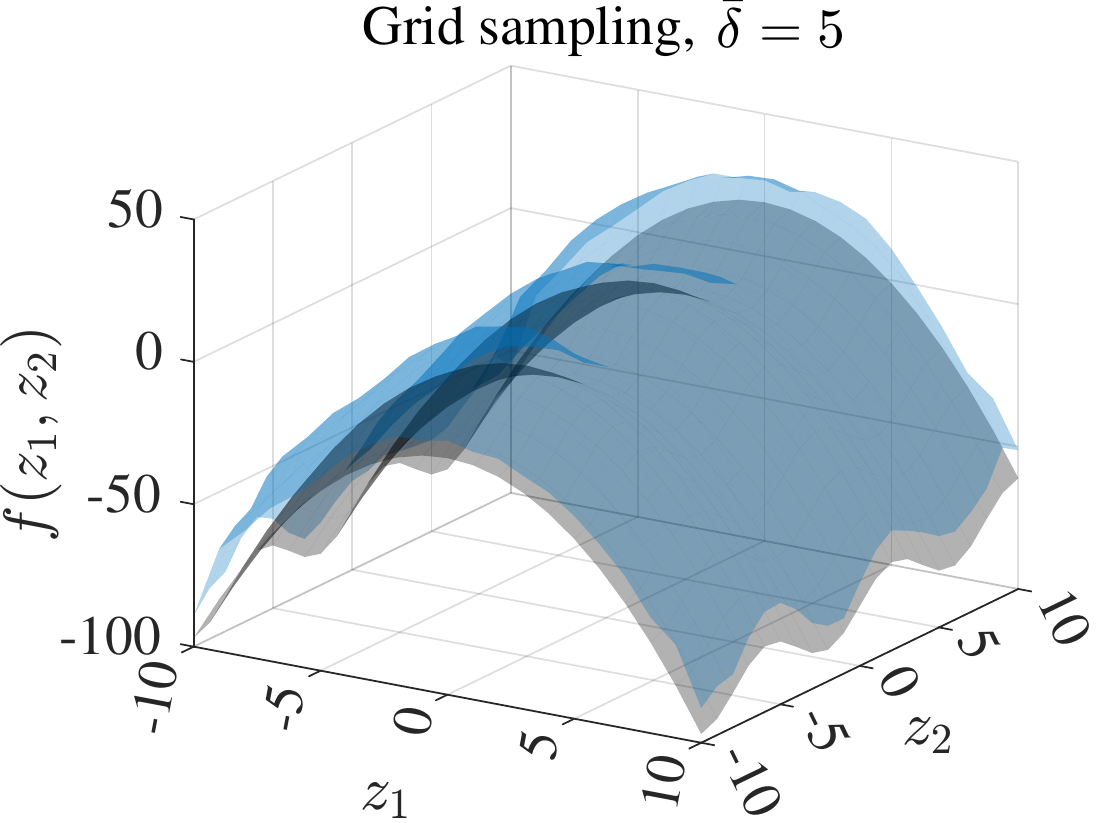}
	\includegraphics[scale=0.38]{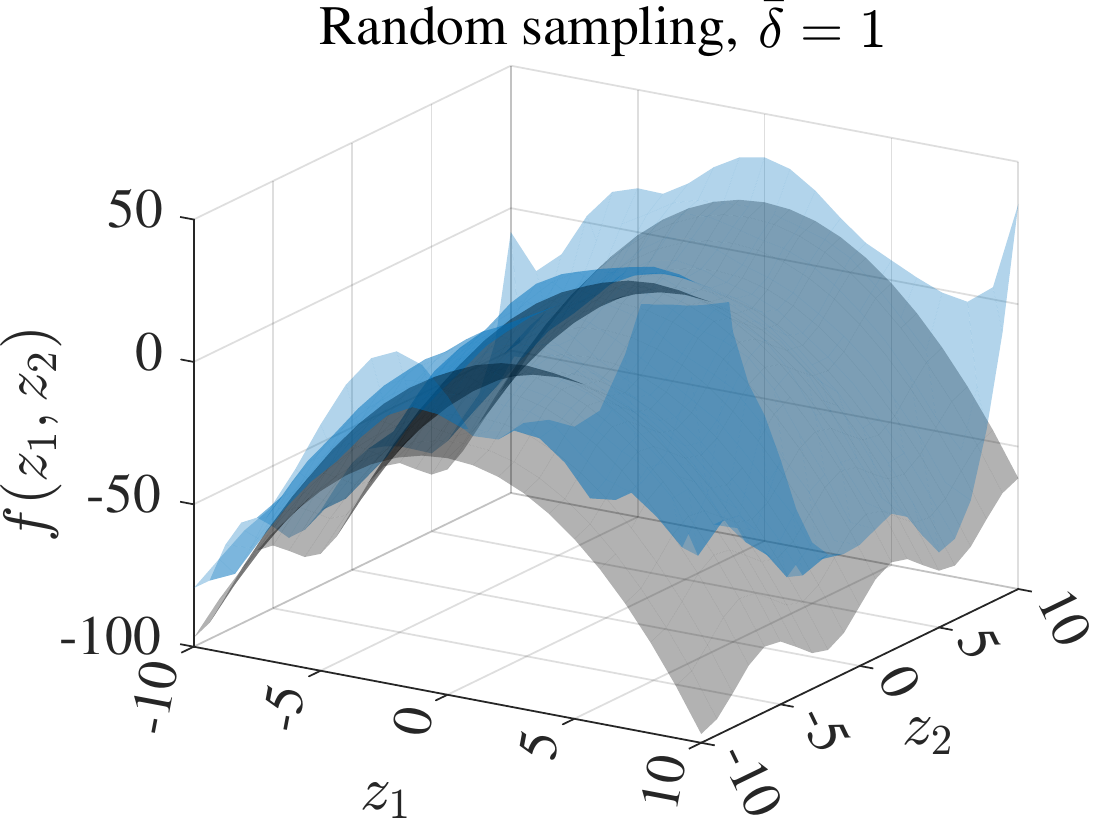}
	\includegraphics[scale=0.38]{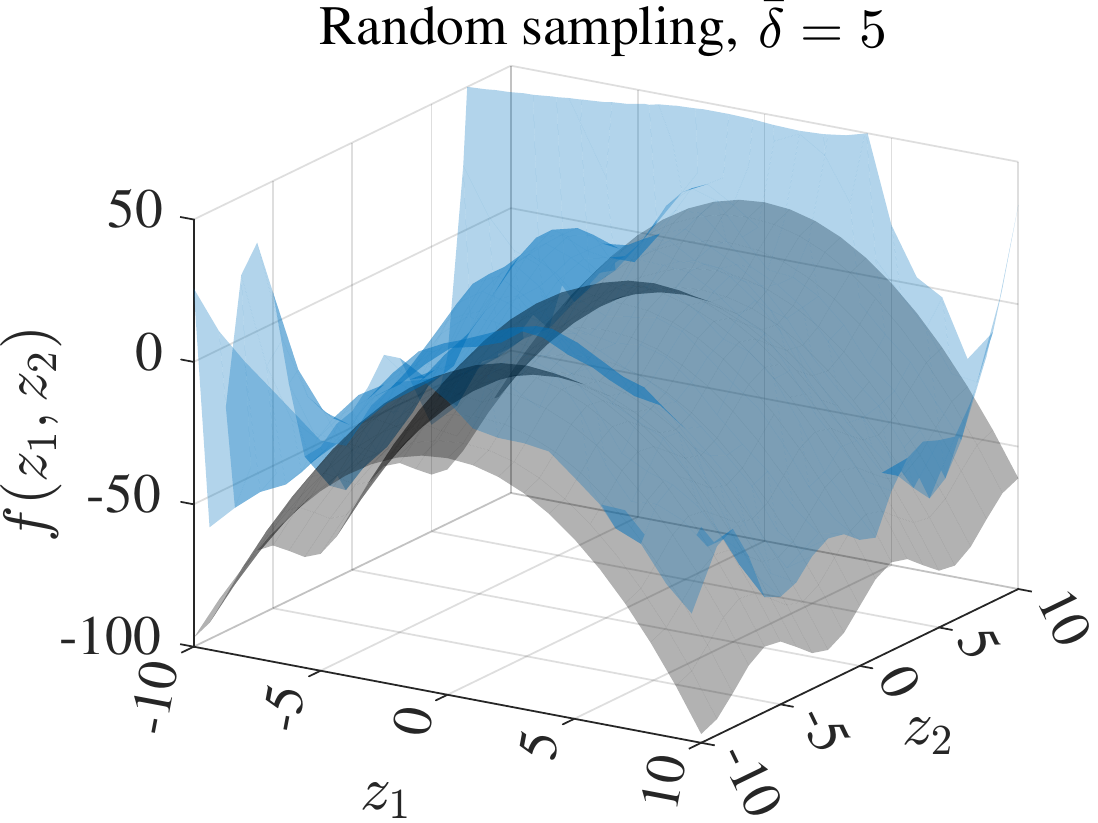}
	\caption{The ground-truth (black) and the upper optimal bound C$(x)$ (blue) with $100$ data-points. Two noise levels are considered, $\bar\delta = 1$ and $\bar\delta = 5$, and two sampling strategies, an equidistant grid and random uniform sampling.}
	\label{fig:gridrandcomp}
\end{figure*}
\begin{figure}[htbp!]
	\includegraphics[scale=0.38]{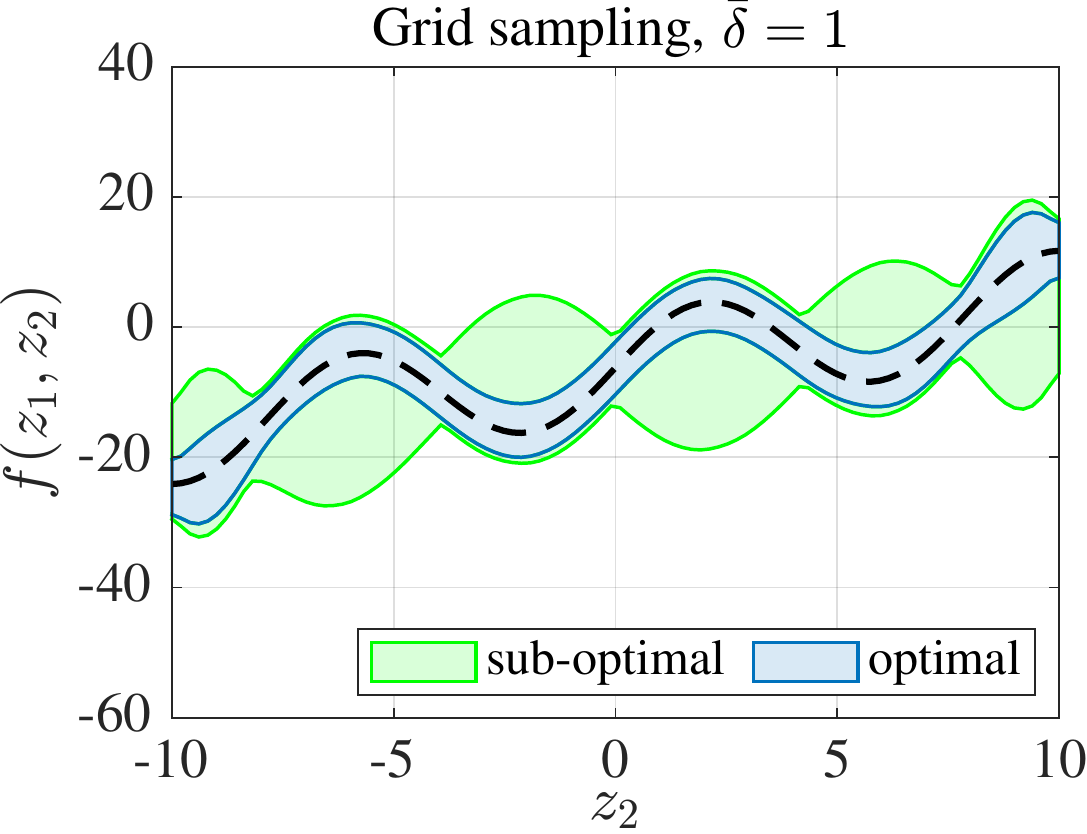}
	\includegraphics[scale=0.38]{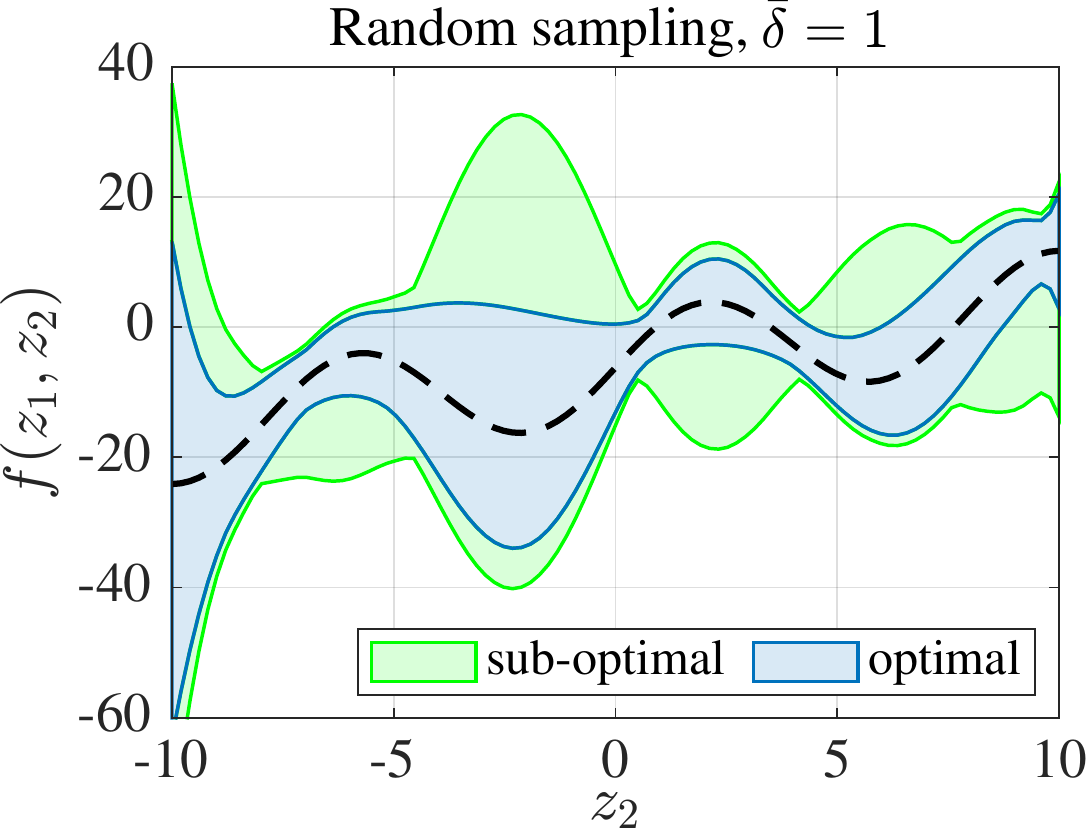}
	\caption{A comparison between the optimal bounds (blue) and the closed-form sub-optimal ones (green). The 2D ground-truth was sliced at $z_1 = -3$ and is shown in dashed black.}
	\label{fig:krrbd}
\end{figure}

\section{Numerical examples}

\red{The methods developed in the previous sections are now employed in three distinct scenarios: a function bounding task, an optimization problem with an unknown constraint, and a control certification procedure.\footnote{The code to reproduce our results is available at \texttt{\url{https://github.com/PREDICT-EPFL/opt-rkhs-bounds}}.}}

\label{sec::Result}
\textbf{Example 1:} Consider the function below, which represents the first component update map of a Hénon chaotic attractor with an additional sinusoidal forcing term
\begin{equation}
	\label{eq:gtfunc}
	f^\star(z_1,z_2) = 1 - az_1^2 + z_2 + b\sin(c\,z_2)
\end{equation}
The parameters are $a=0.8$, $b=8$ and $c=0.8$, and its domain is the box $\Omega = \begin{bmatrix}
-10 & 10 \end{bmatrix} \times \begin{bmatrix}
-10 & 10 \end{bmatrix}$. A squared-exponential kernel $k(x,x') = \exp \left(\frac{\Vert x - x'\Vert^2}{2\ell^2}\right)$ with $x = \begin{bmatrix} z_1 & z_2 \end{bmatrix}$ was chosen for our experiments with lengthscale $l=5$, which was empirically estimated by gridding the search-space and performing posterior validation. $\Gamma$ was obtained through the procedure described in Appendix~\ref{app.normEst} with a final value of $\Gamma=1200$. $d=100$ samples were collected using two strategies: inputs lying in an equidistant grid, and inputs being drawn randomly from a uniform distribution. Noise was sampled uniformly throughout the tests with $\bar \delta = 1$ and $\bar \delta = 5$.

The obtained optimal upper bound $\text{C}(x)$ is displayed in Figure~\ref{fig:gridrandcomp} along with the ground-\red{truth} function $f^\star$. Consider the \red{scenarios where} $\bar \delta = 1$. Whereas the $\text{C}(x)$ surface is overall tight for the grid-based dataset, with an average distance of $3.01$ to the latent function, randomized data yielded a less regular bound with an average distance of $8.02$. These numbers were increased respectively to $9.57$ and $18.97$ when the noise levels were risen to $\bar \delta = 5$. The plots illustrate the disadvantages of relying on completely randomized input locations, which degraded especially the borders of $\text{C}(x)$. An equidistant grid of points is highly favorable since it not only fills the domain well, but also ensures a minimum separation distance so that no two inputs are too close to cause numerical problems when handling the kernel matrix $K_{XX}$. 

The $f^\star(z_1,z_2)$ map was then sliced at $z_1 = -3$ and the entire envelope $\text{B}(x) \leq x \leq \text{C}(x)$ was computed. This was compared to the sub-optimal bounds given in Proposition~\ref{prop:uniform} for a kernel ridge regression (KRR) model. The two previous datasets with $\bar \delta = 1$ were used and the obtained results are \red{displayed} in Figure~\ref{fig:krrbd}. As can be seen from the plots, \red{the optimal approach yielded tighter uncertainty intervals that were always within the sub-optimal ones. Moreover, whereas the average width of the blue envelope was 8.93 and 18.96 respectively in the grid and random cases, the green envelope displayed average widths of 21.13 and 34.62.}

\textcolor{black}{Next, we consider the Gaussian process bounds proposed in \cite[Lemma~3]{berkenkamp2017safe} (see also the closely related works \cite{koller_learning_2018,chowdhury2017kernelized}) and analyze how they compare to the proposed robust ones. Overloading notation for the sake of clarity, these bounds have the form
\begin{subequations}
	\label{eq:gpbounds}
	\begin{align}
        &|\mu(x)-f^\star(x)| \leq \beta \, \sigma(x) \\
		\text{with }\;\;&\beta = \Gamma + 4 \lambda \sqrt{\gamma + 1 + \ln(1/\delta)},
	\end{align}
\end{subequations}
where $\mu(x)$ is the GP mean, $\sigma(x)$ is its standard deviation, $\lambda$ is the strength of the sub-Gaussian noise, $\gamma$ is the maximum information capacity for a fixed number of samples, and $1-\delta$ is the confidence of the inequality. For a detailed explanation of how the experiment was set up, the reader is referred to Appendix~\ref{app:gpcomp}. The data, $d=100$ samples, corrupted by the same noise realizations were used throughout the tests for all methods. Two parameters were then varied to understand how sensitive each method is to them: the RKHS norm estimate $\Gamma$ and the noise bound $\bar \delta$, which were increased by a factor of 1, 1.5, and 2. Detailed results can be found in Appendix~\ref{app:gpcomp}, Tables~\ref{tab:compmodnoise} and \ref{tab:comphighnoise}. The outcomes in all 18 different scenarios were unanimous in ranking the optimal bounds as the tightest method, followed by the sub-optimal ones, and finally the GP approach. Indeed, the GP bounds always yielded average widths at least one order of magnitude greater than the optimal deterministic ones. We attribute this difference especially to the direct product between of $\Gamma$ and $\sigma(x)$ in \eqref{eq:gpbounds}, which causes them to be particularly sensitive to norm over-approximations. This effect is dampened in \eqref{eq.uniformBound2} due to the interaction with $\tilde\Delta$ (see the derivation in Appendix~\ref{app.lemproof}).}

\begin{figure*}[t]
	\centering
	\includegraphics[scale=0.25]{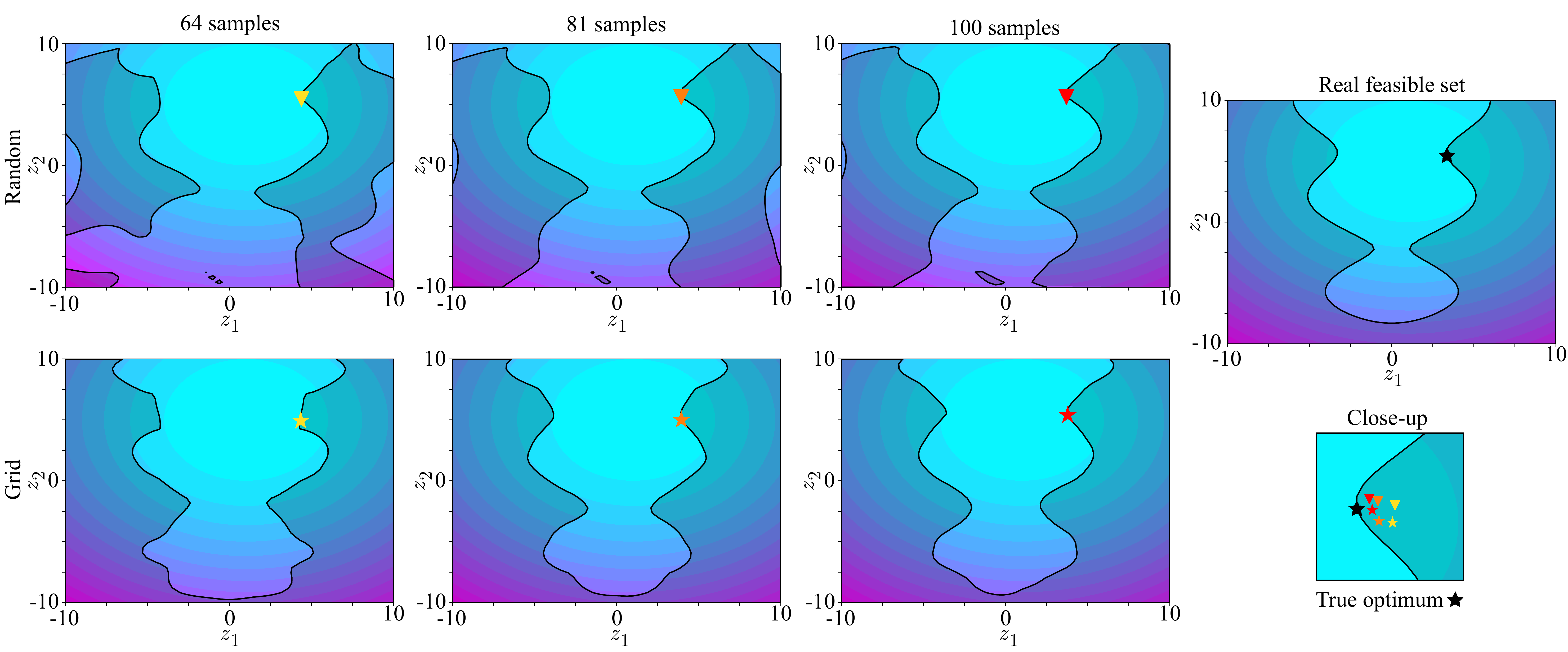}
	\caption{Solutions and feasible sets (shaded areas) for problem \eqref{eq:rto} with 64, 81 and 100 samples of $f^\star(z)$. Top row: samples drawn uniformly. Bottom row: samples on an equidistant grid. The true feasible set and optimal solution are shown on the right.}
	\label{fig:constraints}
\end{figure*}

\textbf{Example 2:} The next numerical experiment involves the ground-truth \eqref{eq:gtfunc} as an unknown constraint for a static problem (data-driven optimization with unknown constraints is \red{typical} in the field of real-time optimization \cite{chachuat2009adaptation}). Consider the following formulation
\begin{subequations}
	\label{eq:rto}
	\begin{align}
		\min_{z \in \mathbb{R}^2} & \quad (z_1-1)^2 + (z_2-5)^2  \\ 
		\text{\normalfont subj. to} & \ \; 
		f^\star(z) \le -10 \label{eq:rtocons}
	\end{align}
\end{subequations}
\red{where the function $f^\star(z)$ that maps the decision variables to the constraint is not explicitly known, but can be measured}. Samples were used to establish an upper bound $\text{C}(z)$ for $f^\star(z)$, hence providing an inner-approximation for the real feasible set. We considered the cases of having 64, 81 and 100 evaluations of $f^\star(z)$ affected by noise with $\bar\delta=1$ and, once more, the data were collected by means of a uniform random distribution and an equidistant grid. In the approximate optimization problems, the original constraint \eqref{eq:rtocons} was replaced by $C(z)\le -10$. Optimizers $z^\star$ were computed by gridding the domain, and the results along with the estimated feasible sets (shaded areas) are shown in Figure~\ref{fig:constraints}. Notice how in some instances the set of feasible decisions is not connected. Thanks to Proposition~\ref{prop.decreasing}, the addition of new data-points can only relax the approximate formulation, hence reducing the found minimum.
\red{Indeed, the obtained solutions for the approximate problems were $13.21$, $11.36$ and $10.96$, respectively with $64$, $81$ and $100$ samples taken randomly. When employing a grid, the figures were $10.67$, $8.48$ and $7.67$. The solution of the real problem, i.e., the one with the ground-truth constraint, is $5.69$.}

\red{
\textbf{Example 3:} Finally, we verify if a sequence of control actions obtained by means of a certainty equivalence approach will or will not lead to the real system violating constraints. In this scenario, previous examples in the form of control and state trajectories are exploited to build the necessary datasets.

Let us consider a continuous stirred-tank reactor (CSTR) described by the differential equations \begin{subequations}
	\begin{align}
		\dot{c}_A(t) &= u(t)(c_{A0}-c_A(t)) - \rho_1 c_A(t) - \rho_3 c_A^2(t)  \\ 
		\dot{c}_B(t) &= -u(t)c_B(t) + \rho_1 c_A(t) - \rho_2 c_B^2(t)
	\end{align}
\end{subequations}
$c_A$ and $c_B$ denote respectively the concentrations of cyclopentadiene and cyclopentenol, whereas $u$ represents the feed inflow of cyclopentadiene. The parameters are $\rho_1 = \rho_2 = 4.1 \times 10^{-3}$ h$^{-1}$, $\rho_3 = 6.3 \times 10^{-4}$ h$^{-1}$, $c_{A0} = 5.1$ mol/l. The system is subject to the constraints $1 \leq c_A \leq 3$, $0.5 \leq c_B \leq 2$, $3 \leq u \leq 25$, and is sampled at a rate of $1/30~$Hz. In order to steer the CSTR states toward $c_A^{\text{ref}} = 2.14$, $c_B^{\text{ref}} = 1.09$, an optimal control problem (OCP) based on KRR models was formulated and solved. The approach featured no uncertainty quantification, i.e., it relied solely on certainty equivalence. 

To verify if OCP control actions would not lead to the true system violating constraints, the tools developed in Section~\ref{sec::OptimalBound} were employed. The certification problem was broken down into several steps: the 1-step ahead analysis, the 2-step ahead analysis, and so on. The associated datasets $\{(x_i,\mathsf{y}_i)\}_{i=1}^d$ were composed of initial states and sequences of control actions to form $x_i$, and the final state to form $\mathsf{y}_i$ (this multi-step approach is the same as the one explained in \cite[Sec.~4]{maddalena2020kpc}). The squared-exponential kernel was used throughout the whole process and the various lengthscales were selected through a 5-fold cross-validation process based on 400 samples. The same batch of data were exploited to estimate the different RKHS norms $\Gamma$ following the procedure of Appendix~\ref{app.normEst}. The obtained lower estimates were then augmented to account for possible unseen complexity. A different dataset was gathered to compute the bounds by starting the system at $800$ initial conditions and solving OCP from those locations. We highlight that, as there are two states and one control variable, the domain of the ground-truth mapping the initial condition to the 8-th step ahead state has dimension 10, hence justifying the need for a large dataset. The noise affecting the measurements was drawn uniformly from the interval $-0.05$ to $0.05$, and a bound of $\bar \delta = 0.06$ was used.}

\begin{figure}[htbp!]
	\includegraphics[scale=0.38]{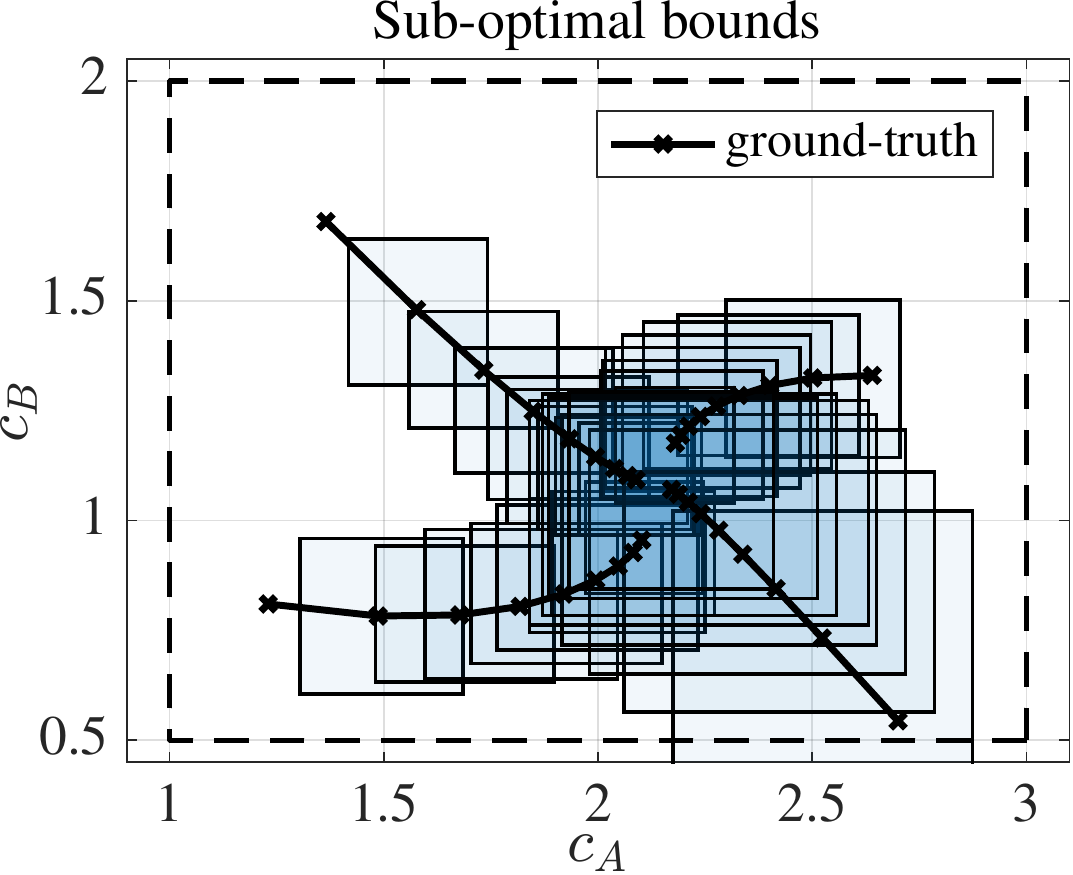}
	\includegraphics[scale=0.38]{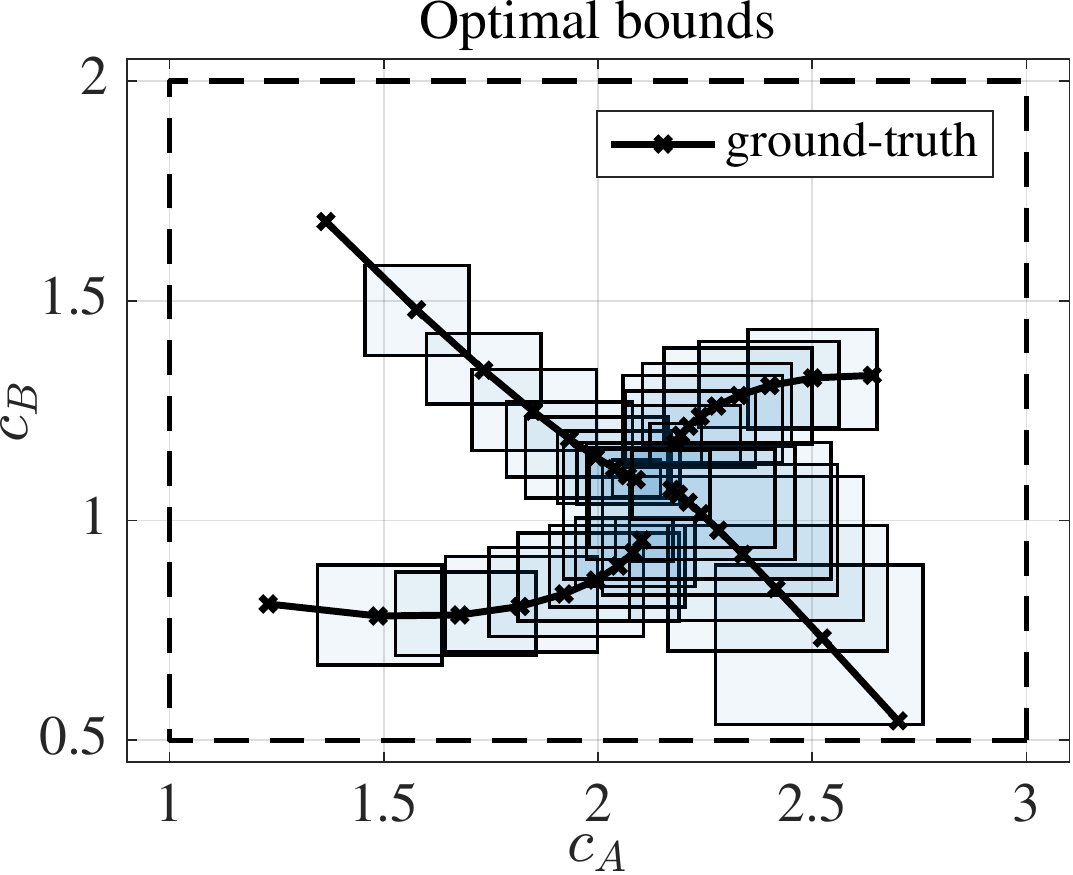} \hspace{1pt}
	\caption{\red{Phase portraits of the CSTR system under the same control inputs, but with different uncertainty quantification techniques. Constraints are represented by the dashed lines.}}
	\label{fig:cstr_phase}
	\vspace{8pt}
    \includegraphics[scale=0.38]{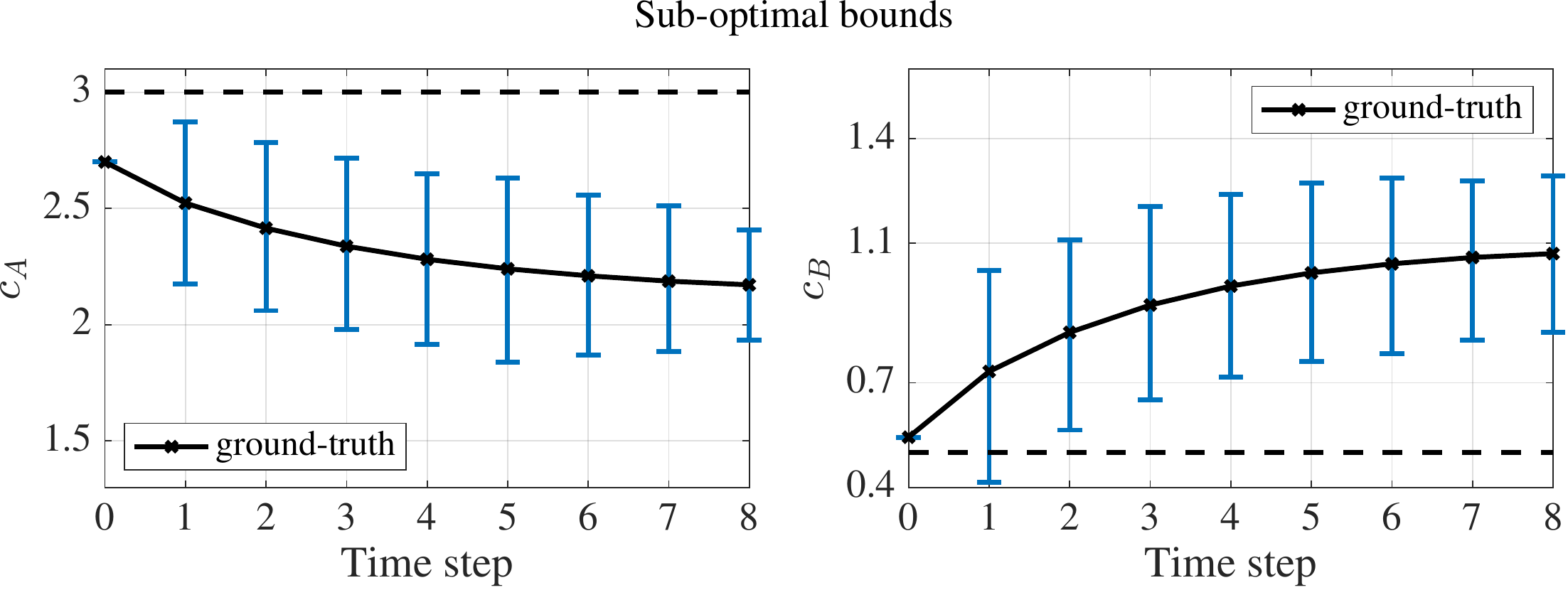} \hspace{3pt}\\
    \includegraphics[scale=0.38]{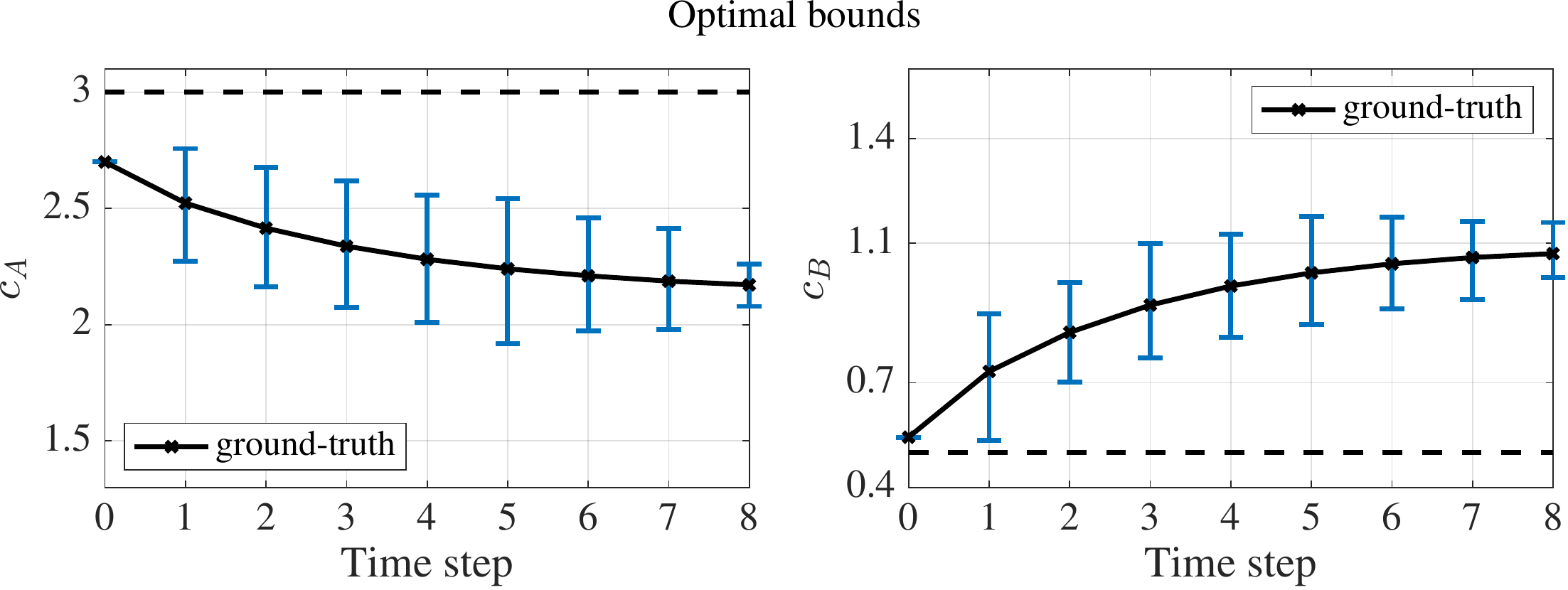} \hspace{3pt}
	\caption{\red{State trajectories of the CSTR system under the same control inputs, but with different uncertainty quantification techniques. Constraints are represented by the dashed lines.}}
	\label{fig:cstr_time}
\end{figure}

\red{The two types of bounds were then computed defining an interval per state and, thus, a ``bounding box" for each step. These are then guaranteed to contain the true system evolution, the ground-truth, as illustrated in Figure~\ref{fig:cstr_phase}. 
After visually inspecting the phase portraits, one sees how conservative the sub-optimal method was: the average area of the sub-optimal boxes was $0.1780$, and $0.0741$ in the optimal case. 
In addition to it, one bounding box around the trajectory that starts at the bottom right corner of the plots extends below the $c_B \geq 0.5$ constraint. A time-domain view of the situation is shown in Figure~\ref{fig:cstr_time}, where at time-step $1$ the lower-bound violates the aforementioned constraint in the top plot, but not in the bottom one. As a result, the OCP control sequence could not be certified by the sub-optimal approach, but could by means of the optimal one.}

\section{Final remarks and future directions} 
\label{sec::Conclusion}
We investigated the uncertainty quantification problem associated with evaluations of an unknown function that belongs to a possibly infinite-dimensional reproducing-kernel Hilbert space. Optimal robust bounds were derived by exploiting a finite set of samples and an estimate of the ground-truth function complexity as measured by its norm. Several formulations were then analyzed: a primal finite-dimensional program, one possible dual form, as well as closed-form sub-optimal solutions centered around pre-specified kernel models. When considering the optimal alternatives, it was shown how the addition of new data can only shrink the uncertainty envelope everywhere.

Future research could focus on the following topics. Firstly, the developed theory could be generalized to accept uncertain inputs, thus allowing for uncertainty propagation in multi-stage problems. Additionally, resampling techniques could be used to construct sparse representations or to confer a desired geometrical property on the input points, enabling fast evaluation of the bounds. \textcolor{black}{Exploring further estimation techniques for $\Gamma$ and $\bar \delta$, especially joint estimation, could be of interest for practical application of the approach. }Finally, the developed finite-sample bounds could give support to the area of data-driven optimization under unknown constraints or objectives by establishing formal feasibility or performance guarantees.

\appendices

\section{Estimating RKHS complexity from data}
\label{app.normEst}
\begin{figure}[t!]
	\centering
	\includegraphics[scale=0.38]{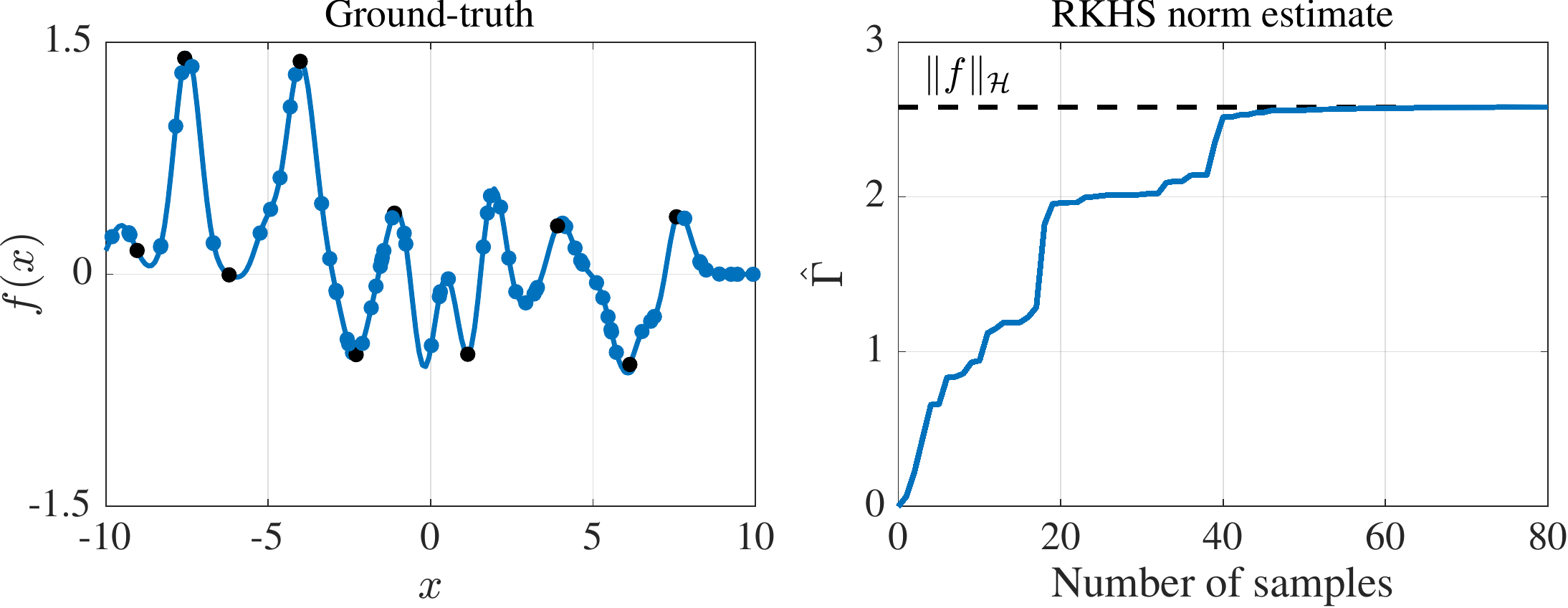} \hspace{1pt}
	\caption{Estimating the RKHS norm using \textcolor{black}{randomly sampled data (all circles). The quadratic form $\hat \Gamma$ for the random samples is shown on the right plot.
	If one sampled only the black subset of the data, the corresponding $\hat \Gamma$ would capture over 90\% of the total complexity, i.e., $\hat \Gamma / \Vert f \Vert_\mathcal{H} > 0.9$.}}
	\label{fig.normEst}
\end{figure}

We consider an unknown map $f\in\mathcal{H}$ and a set of samples $D = \{(x_i,f(x_i))\}_{i=1}^d$. Using the shorthand $f_X = \begin{bmatrix} f(x_1) \dots f(x_d) \end{bmatrix}^\top$, we have that 
\begin{equation}
    \label{eq.gammaHat}
    \hat \Gamma := \sqrt{f_X^\top K_{XX}^{-1} f_X} \leq \rknorm{f}
\end{equation}
for any number of samples $d \in \mathbb{N}$ due to the optimal recovery property \cite{wendland2004scattered}. Moreover, the decomposition used in the proof of Proposition~\ref{prop.decreasing} shows that the quantity $\hat \Gamma$ can only increase with the addition of new data. Since $\rknorm{f}$ is the least upper bound for it, then this quantity can be used as an efficient lower estimate for the RKHS norm. An example is shown in Figure~\ref{fig.normEst} for an $f$ composed of $25$ squared-exponential kernel functions \red{from which 80} samples were drawn uniformly (left). The \red{corresponding values for $\hat \Gamma$ for an increasing number of data are also reported} (right). \red{After around 40 samples, essentially all of the RKHS complexity had already been captured. Moreover, by sampling only the peaks and valleys indicated by the black subset of the data-points, one could retrieve over 90\% of the final norm.} In a practical situation, expert knowledge should be elicited to augment $\hat\Gamma$ \red{through a safety factor} and \red{hopefully} transform it into an upper bound $\Gamma \geq \rknorm{f}$. \red{Note however that no hard guarantees are offered|a situation similar to estimating Lipschitz constants purely from scattered observations.
} Finally, \red{in case} the outputs are corrupted by measurement noise, it is possible to quantify its worst-case effect on the estimation process \cite{maddalena2020deterministic}. 

\section{The data-selection matrix}
\label{app.thebigmatrix}

Recall that $n_1, n_2, \dots, n_d$ are the number of outputs available at the input locations $x_1,x_2,\dots,x_d$. $\Lambda$ has size $(\sum_i n_i) \times d$ and is defined as
\begin{equation}
	\Lambda :=
	\begin{bmatrix}
		\bm{1}_{n_1} & \bm{0}_{n_1} & \bm{0}_{n_1} & \cdots & \bm{0}_{n_1} \\
		\bm{0}_{n_2} & \bm{1}_{n_2} & \bm{0}_{n_2} & \cdots & \bm{0}_{n_2} \\
		\vdots & \vdots & \vdots & \ddots & \vdots \\ 
		\bm{0}_{n_d} & \bm{0}_{n_d} & \bm{0}_{n_d} & \cdots & \bm{1}_{n_d} 
	\end{bmatrix}
\end{equation}
where $\bm{1}_{n_i}$ and $\bm{0}_{n_i}$ are respectively column vectors of ones and zeros of size $n_i$. If only a single output is available at every input, $\Lambda$ simplifies to an identity matrix.

\section{Derivations}
\subsection{Proof of Theorem~\ref{thm.main}}
\label{app.thmproof}

Let $\mathds{X} := X \cup \{x\}$ and define the finite-dimensional subspace $\mathcal{H}^\Vert=\{f\in\mathcal{H}: f \in \text{span}(k(x_i,\cdot), x_i \in \mathds{X})\}$. Furthermore, let $\mathcal{H}^\perp = \{g\in \mathcal{H}: \rkinner{g}{f^\Vert}=0, \forall f^\Vert \in \mathcal{H}^{\Vert} \}$ be the orthogonal complement of $\mathcal{H}^{\Vert}$. Then, we have $\mathcal{H}=\mathcal{H}^{\Vert} \oplus \mathcal{H}^\perp$ and for all $f\in \mathcal{H}$, $\exists f^{\Vert} \in \mathcal{H}^{\Vert}, f^\perp \in \mathcal{H}^\perp : f = f^{\Vert} + f^\perp$. By employing the latter decomposition and using the reproducing property, we can reformulate $\mathds{P}0$ in terms of $\mathcal{H}^{\Vert}$ and $\mathcal{H}^\perp$ as
\begin{align}
    & \sup_{\substack{f^{\Vert} \, \in \mathcal{H}^{\Vert} \\ f^\perp \, \in \mathcal{H}^\perp}} \hspace{-12pt} &&\left\{ 
    \begin{aligned}
    &\rkinner{f^{\Vert}+f^\perp}{k(x,\cdot)} :\\ &\rknorm{f^{\Vert}+f^\perp}^2 \leq \Gamma^2, \infnorm{(f^{\Vert}+f^\perp)_X - \mathsf{y}} \leq \bar \delta 
    \end{aligned}
    \right\} \\
	\overset{(i)}{=}& \sup_{\substack{f^{\Vert} \, \in \mathcal{H}^{\Vert} \\ f^\perp \, \in \mathcal{H}^\perp}} \hspace{-25pt} &&\left\{ f^{\Vert}(x) : \rknorm{f^{\Vert}}^2+\rknorm{f^\perp}^2 \leq \Gamma^2, \infnorm{f_{X}^\Vert - \mathsf{y}} \leq \bar \delta \right\} \\
	\overset{(ii)}{=}& \sup_{f^{\Vert} \, \in \mathcal{H}^{\Vert}} \hspace{-23pt} &&\left\{ f^{\Vert}(x) : \rknorm{f^{\Vert}}^2 \leq \Gamma^2, \infnorm{f^{\Vert}_{X} - \mathsf{y}} \leq \bar \delta \right\} \label{eq.last}
\end{align}
In $(i)$, the $f^\perp$ component vanished from the cost and from the last constraint due to orthogonality w.r.t. $k(x_i,\cdot) \in \mathcal{H}^\Vert$ for any $x_i \in \mathds{X}$; moreover, the Pythagorean relation $\rknorm{f}^2 = \rknorm{f^{\Vert}}^2 + \rknorm{f^\perp}^2$ was also used. To arrive at the second equality $(ii)$, one only has to note that the objective is insensitive to $f^\perp$ and that any $f^\perp \neq 0_{\mathcal{H}}$ would tighten the first constraint. 

The attainment of the supremum is addressed next. Consider \eqref{eq.last} and denote the members of $\mathcal{H}^\Vert$ simply as $f$. $\rknorm{f}^2 \leq \Gamma^2$ is a closed and bounded constraint as it is the sublevel set of a norm. We transform $\infnorm{f_{X} - \mathsf{y}} \leq \bar \delta$ into $\vert f(x_i) - y_{i,j} \vert \leq \bar \delta$, $i=1,\dots,d, \, j =1,\dots,n_i$. Sets of the form $\{a \in \mathbb{R}: \vert a \vert \leq b \}$ are clearly closed in $\mathbb{R}$, hence \{$f(x_i) \in \mathbb{R} : \vert f(x_i) - y_{i,j}\vert \leq \bar \delta, \forall i,j \}$ is also closed. For any $x_i$, the evaluation functional $L_{x_i}(f) = f(x_i)$ is a linear operator and thus pre-images of closed sets are also closed. Consequently, $\{f \in \mathcal{H}^\Vert : \vert f(x_i) - y_{i,j}\vert \leq \bar \delta, \forall i,j \}$ is closed in $\mathcal{H}^\Vert$. The intersection of a finite number of closed sets is necessarily closed, thus all constraint present in \eqref{eq.last} define a closed feasible set. Since $\mathcal{H}^{\Vert}$ is finite-dimensional, any closed and bounded subset of it is compact (Heine–Borel); therefore, the continuous objective $L_x(f) = f(x)$ in \eqref{eq.last} attains a maximum by the Weierstrass extreme value theorem. 

Finally, we establish the connection between $\mathds{P}0$ and $\mathds{P}1$. From the above arguments, an optimizer for $\mathds{P}0$ must lie in $\mathcal{H}^\Vert$. The members $f \in \mathcal{H}^\Vert$ have the form $f(z) = \alpha^\top K_{\mathds{X}z}$, being defined by the $\alpha$ weights. Due to the positive-definit\textcolor{black}{e}ness of $k$, there exists a bijective map between outputs at the $\mathds{X}$ locations $f_{\mathds{X}} = \begin{bmatrix} f(x_1) & \dots & f(x_d) & f(x) \end{bmatrix}^\top$ and the weights $\alpha$, namely $\alpha = \textcolor{black}{K_{\mathds{X}\mathds{X}}^{-1} }f_{\mathds{X}}$. $\textcolor{black}{K_{\mathds{X}\mathds{X}}}$ denotes the kernel matrix associated with the set $\mathds{X} = X \cup \{x\}$. Consequently, optimizing over $f \in \mathcal{H}^\Vert$ is equivalent to optimizing over $\begin{bmatrix} f(x_1) & \dots & f(x_d) & f(x) \end{bmatrix}^\top =: \begin{bmatrix}c^\top & c_x\end{bmatrix}^\top$. The bounded norm condition can be recast as $\rknorm{f}^2 = \rkinner{f}{f} = \alpha^\top \textcolor{black}{K_{\mathds{X}\mathds{X}}} \alpha = \begin{bmatrix}c^\top & c_x\end{bmatrix} \textcolor{black}{K_{\mathds{X}\mathds{X}}^{-1}} \begin{bmatrix}c^\top & c_x\end{bmatrix}^\top$. The remaining constraint and the objective are straightforward. Noting that this reformulation is valid for any $x \in \Omega$ concludes the proof. \QED

\subsection{Proof of Proposition~\ref{prop:uniform}}
\label{app.lemproof}
For any given $s(x)=\alpha^\top K_{Xx}$, we have 
\begin{align}\notag
	&\vert f^\star(x) -s(x)\vert \\=& \, \vert f^\star(x) -\tilde{s}(x) + \tilde{s}(x)-s(x)\vert \\
	\le& \, \vert f^\star(x) - (f^\star_X + \delta_X) K_{XX}^{-1}K_{Xx}\vert + \vert  \tilde{s}(x)-s(x)\vert \label{eq:triag}\\
	\le& \, \vert f^\star(x) - \bar s(x)\vert + \bar \delta \onenorm{K_{XX}^{-1}K_{Xx}} + \vert  \tilde{s}(x)-s(x)\vert \label{eq:triag2}\\
	\le& \, P(x) \, \sqrt{ \Gamma^2  - \rknorm{\bar{s}}^2} + \bar \delta \onenorm{K_{XX}^{-1}K_{Xx}} + \vert  \tilde{s}(x)-s(x)\vert \label{eq:bdnoisefree}\\
	\le& \, P(x) \, \sqrt{ \Gamma^2 + \Delta - \rknorm{\tilde{s}}^2} + \bar \delta \onenorm{K_{XX}^{-1}K_{Xx}} + \vert  \tilde{s}(x)-s(x)\vert \label{eq:Deltdef}
\end{align}
with $f^\star_X$ being the vector of true function values at the sample locations in $X$ and $\delta_X$ the vector of additive measurement noise for the samples $y$. \eqref{eq:triag} follows from the triangle inequality and the additive noise property of $y$. Using the triangle inequality again, we arrive at \eqref{eq:triag2}, where $\bar s$ denotes the noise-free interpolant of $f^\star_X$. The noise-free interpolation error bound gives the estimation in the first term of \eqref{eq:bdnoisefree}, while \eqref{eq:Deltdef} follows from \cite[Lemma~1]{maddalena2020deterministic}, with $\Delta = \max_{\infnorm{\delta} \leq \bar{\delta}} (- \delta^\top K_{XX}^{-1} \delta + 2 y ^\top K_{XX}^{-1} \delta)$. A standard dualization procedure as the one presented in Appendix~\ref{app.dual} leads to the dual problem
\begin{equation}
	\label{eq:dualdelt}
	\min_{\nu \in \mathbb{R}^d} \frac{1}{4} \nu^\top K_{XX} \nu + \nu^\top y + \bar{\delta} \onenorm{\nu} + y^\top K_{XX}^{-1} y
\end{equation}
for $\Delta$. Notice that the last term in \eqref{eq:dualdelt} is constant and the same as the squared interpolant norm $\rknorm{\tilde{s}}^2$. Therefore, these terms cancel in \eqref{eq:Deltdef} and we are left with
\begin{equation}
\begin{aligned}
\vert f^\star(x)-s(x)\vert  \le \,& P(x) \, \sqrt{ \Gamma^2 + \tilde\Delta} + \bar \delta \onenorm{K_{XX}^{-1}K_{Xx}} \\
&+ \vert  \tilde{s}(x)-s(x)\vert
\end{aligned}
\end{equation}
where $\tilde{\Delta}$ represents \eqref{eq:dualdelt} without the constant term.

\subsection{The Lagrange dual problem}
\label{app.dual}

Consider the case $x \not\in X$. Let $z := \begin{bmatrix} c^\top & c_x\end{bmatrix}^\top$, $a := \begin{bmatrix} \textbf{0}^\top & 1\end{bmatrix}^\top$, $A := \begin{bmatrix} \textbf{I} & \textbf{0}\end{bmatrix}$. The Lagrangian of $\mathds{P}1$ is
\begin{align}\label{eq.lagrangianP2}
    \mathcal{L}(z,\lambda,\beta,\gamma) =& a^\top z - \lambda (z^\top \textcolor{black}{K_{\mathds{X}\mathds{X}}}^{-1}z - \Gamma^2) \\\notag
    &- \beta^\top(\Lambda Az - \mathsf{y} - \bar \delta \textbf{1})- \gamma^\top(\mathsf{y} -\Lambda Az - \bar \delta \textbf{1})
\end{align}
where \textcolor{black}{$K_{\mathds{X}\mathds{X}}$} denotes the kernel matrix evaluated at $X \cup \{x\}$. Suppose $\lambda > 0$. Computing $\nabla_z\mathcal{L}(z^\star) = 0$ leads to
$$z^\star = -\frac{1}{2\lambda}\textcolor{black}{K_{\mathds{X}\mathds{X}}} \left( A^\top \Lambda^\top (\beta - \gamma) - a \right).$$ Defining the auxiliary variable $\nu = \beta - \gamma$, and substituting $z^\star$ into \eqref{eq.lagrangianP2} gives the dual objective
\begin{align}\notag
	g(\lambda,\nu) 
	=\;& \frac{1}{4\lambda} \nu^\top \Lambda A \textcolor{black}{K_{\mathds{X}\mathds{X}}} A^\top \Lambda^\top \nu + \left(\mathsf{y} - \frac{1}{2\lambda} \Lambda A \textcolor{black}{K_{\mathds{X}\mathds{X}}} a \right)^\top \nu \\
	&+ \bar \delta \Vert \nu \Vert_1 +  \frac{1}{4\lambda} a^\top \textcolor{black}{K_{\mathds{X}\mathds{X}}} a + \lambda \Gamma^2 \\\notag
	=\;& \frac{1}{4\lambda} \nu^\top \Lambda K_{XX} \Lambda^\top \nu + \left(\mathsf{y} - \frac{1}{2\lambda} \Lambda K_{Xx} \right)^\top \nu \\
	&+ \bar \delta \Vert \nu \Vert_1 +  \frac{1}{4\lambda} k(x,x) + \lambda \Gamma^2 \label{eq.dualDerivation}
\end{align}
where in the second equality the matrix \textcolor{black}{$K_{\mathds{X}\mathds{X}}$} was expanded and the resulting terms were reorganized. Since $\beta,\gamma \in \mathbb{R}^{\tilde d}_{\geq 0}$ and $\nu = \beta - \gamma$ then $\nu$ is unconstrained.

Now if $\lambda = 0$, the Lagrangian \eqref{eq.lagrangianP2} simplifies to
$\mathcal{L}(z,\nu) = (a - A^\top \Lambda^\top \nu)^\top z + \nu^\top \mathsf{y} + \bar\delta \Vert \nu \Vert_1 \label{eq.lamZero}$, which is linear in $z$. Its supremum w.r.t. $z$ is only finite if $a = A^\top\Lambda^\top\nu$. Recalling the definitions of $a$, $A$ and $\Lambda$, one can see that $\nexists \nu$ that could satisfy the latter condition. Therefore, $\lambda = 0 \implies \sup_z \mathcal{L}(z,\lambda,\nu) = +\infty$, meaning that the dual problem is infeasible. As a conclusion, the Lagrangian dual of $\mathds{P}1$ in \eqref{eq.P2} is precisely $\mathds{D}1$ in \eqref{eq.dualProb}.

Next, consider the case $x \in X$, $x = x_i$. The objective of $\mathds{P}1'$ can be written as $a^\top c$ with $a_i = 1$ and $a_n=0, n \neq i$. When deriving its Lagrangian, one obtains again \eqref{eq.lagrangianP2} with the simplifications: $z \leftarrow c$, $\textcolor{black}{K_{\mathds{X}\mathds{X}}} \leftarrow K_{XX}$ and $A \leftarrow \textbf{I}$. We proceed by analyzing the two scenarios for $\lambda$ as before. If $\lambda > 0$, the previous derivations apply, leading to the same the quadratic-over-linear objective \eqref{eq.dualDerivation}. However, if $\lambda = 0$, the Lagrangian becomes $\mathcal{L}(z,\nu) = (a -  \Lambda^\top \nu)^\top z + \nu^\top \mathsf{y} + \bar\delta \Vert \nu \Vert_1$, whose supremum w.r.t. $z$ is only finite if $a = \Lambda^\top \nu$. In contrast with the previous paragraph, this condition now can be satisfied. It is equivalent to $\nu_{i,1} + \dots + \nu_{i,n_i} = 1$, where the variables are all the multipliers associated with the $i$-th input location $x_i$. The resulting expression can be minimized analytically, yielding the minimum $\min_{j} y_{i,j} + \bar\delta$, i.e., the smallest output available at $x_i$ augmented by the noise bound. Finally, we conclude that the dual objective for $\mathds{P}1'$ is
\begin{equation}
	g(\lambda,\nu) = 
	\begin{cases}
		\eqref{eq.dualDerivation}, & \text{if } \lambda > 0 \\
		\min_j y_{i,j} + \bar\delta, & \text{if } \lambda = 0
	\end{cases}
\end{equation}

As a last observation, a dual problem can also be derived for \eqref{eq.BofX}, calculating the lower part of the envelope. The formulation is analogous to \eqref{eq.dualProb}, assuming the form
\begin{equation}
    \begin{aligned}
        \max_{\nu \in \mathbb{R}^{\tilde d}, \lambda > 0}\;& -\frac{1}{4\lambda} \nu^\top \Lambda K_{XX} \Lambda^\top \nu - \left(\mathsf{y} + \frac{1}{2\lambda} \Lambda K_{Xx} \right)^\top \nu \\
        &- \bar \delta \Vert \nu \Vert_1 - \frac{1}{4\lambda} k(x,x) - \lambda \Gamma^2
    \end{aligned}
\end{equation}
Note that these are distinct objectives, not merely opposites. Therefore, two problems have to be solved to fully quantify the ground-truth uncertainty.

\section{A block matrix identity}
\label{app.blockMatrixIdent}

Let $A \in \mathbb{R}^{d\times d}$ be invertible, $B \in \mathbb{R}^d$ and $c \in \mathbb{R}$. The following identity holds
\begin{equation}
	\begin{bmatrix}
		A & B \\
		B^\top & c
	\end{bmatrix}^{-1} 
	= 	\begin{bmatrix}
		A^{-1} + \frac{1}{d}A^{-1}BB^\top A^{-1} & - \frac{1}{d}A^{-1}B \\[3pt]
		- \frac{1}{d}B^\top A^{-1} & \frac{1}{d}
	\end{bmatrix}
\end{equation}
where $d = c - B^\top A^{-1} B$.

\textcolor{black}{\section{GP comparison: settings and results}
\label{app:gpcomp}}

\textcolor{black}{In order to compare the GP uncertainty bounds \eqref{eq:gpbounds} to their deterministic counterparts, the following approach was adopted. First, a lower bound for the maximum information gain $\gamma$ was used since the problem of exactly computing such a quantity is in general NP-hard \cite{srinivas2012information}. Note how this decision favors the GP bounds by shrinking them. The chosen lower bound was the information gain of our inputs $X$, which in our zero-mean Gaussian noise setting with variance $\lambda^2$ is $\frac{1}{2}\ln( \det (I+\lambda^{-2}K_{XX}))$ \cite{srinivas2012information}. As for the noise realizations, we proceeded as follows. Starting from our hard noise limit $\bar\delta$, we considered a zero-mean Gaussian distribution with variance such that its samples would lie in the $[-\bar\delta, \bar\delta]$ band with confidence $0.99$, i.e., a standard deviation of $\lambda=\frac{\bar\delta}{2.58}$. The noise was then drawn from the normal distribution and clipped to the interval $[-\bar\delta, \bar \delta]$ to fulfill Assumption~\ref{as:noisebound}. Finally, the probabilistic inequality \eqref{eq:gpbounds} was evaluated for a final confidence of 99\%. The obtained numerical results are shown in Tables~I and II.
}

\begin{table*}[t!]
    \centering
    \textcolor{black}{\caption{Average distance between the upper and lower bounds for the optimal (opt) and sub-optimal (sub) deterministic cases, and the Gaussian process alternative (gp). Moderate noise level (true $\bar\delta=1$), using factors of 1, 1.5, and 2 to augment $\bar\delta$ and $\Gamma$.}
\label{tab:compmodnoise}
\begin{tabular}{| c c || c | c | c || c | c | c || c | c | c ||} 
 \hline
 \multicolumn{2}{| r ||}{$\Gamma$} & \multicolumn{3}{c ||}{1200}  & \multicolumn{3}{c ||}{1800} & \multicolumn{3}{c ||}{2400}\\
 \hline
 \multicolumn{2}{| r||}{$\bar\delta$} & 1 & 1.5 & 2  & 1 & 1.5 & 2 & 1 & 1.5 & 2  \\
 \hline \hline
 \multirow{3}{*}{\rotatebox[origin=c]{90}{Grid}} & opt & 6.21 & 8.35 & 10.34 &   7.45  & 9.75 & 11.90 & 8.50 & 10.94 & 13.20  \\
 \cline{2-11}
  & sub & 11.07  & 15.60 & 20.13   & 11.70  & 16.23 & 20.76 & 12.36  & 16.89 & 21.42  \\
  \cline{2-11}
   & gp & 604.51  & 706.13 & 786.51  & 904.61  & 1055.89 & 1175.32 & 1204.71  & 1405.65 & 1564.12  \\
\hline \hline
 \multirow{3}{*}{\rotatebox[origin=c]{90}{Rand}} & opt & 14.62  & 19.02 & 22.89  & 18.05  & 23.08 & 27.51 & 20.85  & 26.39 &  31.26 \\
 \cline{2-11}
  & sub & 64.78  & 93.99 & 123.20  & 65.91  & 95.12 & 124.33 & 67.07  & 96.28 & 125.49  \\
  \cline{2-11}
   & gp & 643.20  & 743.44 & 822.24  & 962.51  & 1111.67 & 1228.70 & 1281.82  & 1479.90 & 1635.17  \\
   \hline \hline
\end{tabular}}
\end{table*}

\begin{table*}[t!]
    \centering
    \textcolor{black}{\caption{Average distance between the upper and lower bounds for the optimal (opt) and sub-optimal (sub) deterministic cases, and the Gaussian process alternative (gp). High noise level (true $\bar\delta=5$), using factors of 1, 1.5, and 2 to augment $\bar\delta$ and $\Gamma$.}
\label{tab:comphighnoise}
\begin{tabular}{| c c || c | c | c || c | c | c || c | c | c ||} 
 \hline
 \multicolumn{2}{| r ||}{$\Gamma$} & \multicolumn{3}{c ||}{1200}  & \multicolumn{3}{c ||}{1800} & \multicolumn{3}{c ||}{2400}\\
 \hline
 \multicolumn{2}{| r||}{$\bar\delta$} & 5 & 7.5 & 10  & 5 & 7.5 & 10 & 5 & 7.5 & 10  \\
 \hline \hline
 \multirow{3}{*}{\rotatebox[origin=c]{90}{Grid}} & opt & 20.29 & 28.57 & 36.39 & 22.54  & 31.31 & 39.58 & 24.41 & 33.56 & 42.17  \\
 \cline{2-11}
  & sub & 49.15  & 71.79 & 94.44  & 49.81 & 72.46 & 95.11 & 50.48  & 73.13 & 95.78  \\
  \cline{2-11}
   & gp & 1090.16  & 1247.34 &  1366.96 & 1624.19  & 1854.47 & 2028.24 & 2158.21  & 2461.60 &  2689.52 \\
\hline \hline
 \multirow{3}{*}{\rotatebox[origin=c]{90}{Rand}} & opt & 39.95  & 53.43 & 65.41  & 47.00  & 62.15 & 75.57 & 52.76  & 69.32 & 83.89  \\
 \cline{2-11}
  & sub & 312.44  & 458.51 & 604.57  & 313.61   & 459.68 & 605.74 &  314.79 & 460.85 & 606.91  \\
  \cline{2-11}
   & gp & 1117.01  & 1268.40 & 1383.43  & 1664.18  & 1885.79 & 2052.67 & 2211.36  & 2503.18 & 2721.91  \\
   \hline \hline
\end{tabular}}
\end{table*}

\IEEEpeerreviewmaketitle

\ifCLASSOPTIONcaptionsoff
  \newpage
\fi

\bibliographystyle{IEEEtran}
\bibliography{SafeLearningBib}

\end{document}